\documentclass[journal]{IEEEtran}
\usepackage{amsmath,amsfonts}
\usepackage{algorithmic}
\usepackage{algorithm}
\usepackage{array}
\usepackage[caption=false,font=normalsize,labelfont=sf,textfont=sf]{subfig}
\usepackage{textcomp}
\usepackage{stfloats}
\usepackage{url}
\usepackage{verbatim}
\usepackage{graphicx}
\usepackage{cite}
\usepackage{graphicx}
\usepackage{amsmath}
\usepackage{amssymb}
\usepackage{booktabs}

\usepackage{algorithm}
\usepackage{algorithmic}
\usepackage[normalem]{ulem}
\useunder{\uline}{\ul}{}
\usepackage{booktabs}

\usepackage{multirow}
\usepackage{booktabs} 
\usepackage{color}

\newcommand{\etal} {\textit{et al.}}

\begin{document}

\title{Learning Sub-Pixel Disparity Distribution for Light Field Depth Estimation}

\author{Wentao Chao, Xuechun Wang, Yingqian Wang, Guanghui Wang,~\IEEEmembership{Senior Member,~IEEE}, Fuqing Duan
\thanks{W. Chao, X. Wang, and F. Duan are with the School of
Artificial Intelligence, Beijing Normal University, Beijing 100875, China. (e-mail: chaowentao@mail.bnu.edu.cn; wangxuechun@mail.bnu.edu.cn; fqduan@bnu.edu.cn).}
\thanks{Y. Wang is with the College of Electronic Science and Technology, National University of Defense Technology, Changsha 410073, China. (e-mail: wangyingqian16@nudt.edu.cn).}
\thanks{G. Wang is with the Department of Computer Science, Toronto Metropolitan University, Toronto, ON M5B 2K3, Canada. (e-mail: wangcs@torontomu.ca).}
\thanks{This work is supported by the National Key Research and Development Project Grant, Grant/Award Number: 2018AAA0100802.}
\thanks{Manuscript received May 14, 2023; revised October 8, 2023, accepted November 11. Corresponding author: F. Duan}}



\maketitle

\begin{abstract}
    Light field (LF) depth estimation plays a crucial role in many LF-based applications. Existing LF depth estimation methods consider depth estimation as a regression problem, where a pixel-wise L1 loss is employed to supervise the training process. However, the disparity map is only a sub-space projection (i.e., an expectation) of the disparity distribution, which is essential for models to learn. In this paper, we propose a simple yet effective method to learn the sub-pixel disparity distribution by fully utilizing the power of deep networks, especially for LF of narrow baselines. We construct the cost volume at the sub-pixel level to produce a finer disparity distribution and design an uncertainty-aware focal loss to supervise the predicted disparity distribution toward the ground truth. Extensive experimental results demonstrate the effectiveness of our method.
   Our method significantly outperforms recent state-of-the-art LF depth algorithms on the HCI 4D LF Benchmark in terms of all four accuracy metrics (i.e., BadPix 0.01, BadPix 0.03, BadPix 0.07, and MSE $\times$100). 
   The code and model of the proposed method are available at \url{https://github.com/chaowentao/SubFocal}.
\end{abstract}


\section{Introduction}
\label{sec:intro}

\IEEEPARstart{L}{ight} Field (LF) \cite{ng2005light, peng2020zero} depth estimation is a fundamental task in computer vision with many subsequent applications, such as refocusing  \cite{ng2005light, wang2018selective}, super-resolution \cite{zhang2019residual, jin2020light, cheng2021light,chen2022light, cheng2022spatial,van2023light }, view synthesis  \cite{wu2018light, meng2019high, jin2020deep}, 3D reconstruction \cite{kim2013scene}, and virtual reality  \cite{yu2017light}. The 4D LF records abundant spatial and angular information in the scene, from which the scene depth can be estimated. Based on the characteristics of light field images, traditional methods \cite{tao2013depth, jeon2015accurate, williem2016robust, zhu2017occlusion} and learning-based methods 
\cite{heber2016convolutional, shin2018epinet,  tsai2020attention, chen2021attention, wang2022occlusion, sheng2023lfnat} have been proposed for light field image depth estimation.

\begin{figure}[htb]
  \centering
  \includegraphics[width=0.9\linewidth]{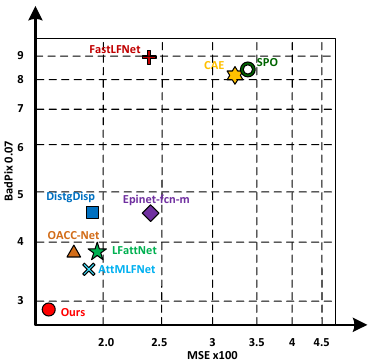}

  \caption{BadPix 0.07 and MSE $\times$100 scores achieved by different methods on the HCI 4D LF benchmark (lower scores represent higher accuracy). }
  \label{fig: intro}
\end{figure}

Early traditional LF depth estimation algorithms can be divided into three categories based on the properties of LF images. Multi-view stereo matching (MVS) based methods \cite{jeon2015accurate} exploit the multi-view information of the LFs for depth estimation. Epipolar plane image (EPI) based methods \cite{zhang2016robust} calculate the slopes using epipolar geometric properties of the LFs to predict the depth. Defocus-based methods \cite{tao2013depth} calculate the depth of a pixel by measuring the consistency at different focal stacks. However, conventional methods involve nonlinear optimization and manually designed features, which are computationally demanding and sensitive to occlusions, weak textures, and highlights. 

Deep learning has been applied to LF depth estimation in recent years  \cite{heber2016convolutional, shin2018epinet, chen2021attention, tsai2020attention, wang2022occlusion}. The mainstream deep learning-based methods typically involve four key steps: feature extraction, cost construction, cost aggregation, and depth (disparity) regression\footnote{Depth $\gamma$ and disparity $d$  can be transformed directly according to $ \gamma = fB/d$, where $B$ and $f$ stand for the baseline length and the focal length of the LF camera, respectively. The paper does not make a distinction between them.}. Cost construction provides initial similarity measures of possible matched pixels among different views.
Existing techniques construct cost volumes by shifting the features of each view with specific integer values, and regress the final disparity values (in sub-pixel accuracy) by performing a weighted average on each candidate disparity. However, the regressed disparity is only a sub-space projection (more precisely, an expectation) of the predicted disparity distribution, which can be considered as a set of probability values that a pixel belongs to different candidate disparities. 
Although these methods have achieved remarkable performance by minimizing the pixel-wise loss between the predicted disparity and the ground truth, the disparity distribution issue has been ignored. 
Since there are unlimited disparity distributions that may produce the same expectation, simply minimizing the loss in the disparity domain can prevent the deep networks from learning the core information and yield sub-optimal depth estimation results.

In this paper, we aim to learn the accurate disparity distribution for LF depth estimation, which requires a sub-pixel cost volume to produce a finer disparity distribution. 
In the previous method \cite{jeon2015accurate}, the sub-pixel cost volume is constructed at the image level by displacing sub-aperture images (SAI) based on the phase shift theorem, which is computationally intensive and time-consuming. In our method, we propose an interpolation-based sub-pixel cost volume construction approach at the feature level to achieve sub-pixel level sampling of disparity distribution. In addition, we design an uncertainty-aware focal loss (UAFL) to reduce the difference between the estimated and ground-truth disparity distributions. Since the ground-truth disparity distribution is unavailable in public LF datasets, we follow the common settings in \cite{li2020generalized} to model the ground-truth disparity distribution as a Dirac delta distribution (i.e., the possibility of the ground-truth disparity is 1 while the others are 0), and further discretize the Dirac delta distribution into two adjacent sampling points to facilitate the training of our method. We conduct extensive experiments to validate the effectiveness of our method. 

In summary, the contributions of this paper are as follows:
\begin{itemize}
\item We reframe LF depth estimation as a pixel-wise distribution alignment problem and make the first attempt to learn the sub-pixel level LF disparity distribution using deep neural networks. 

\item We propose a simple yet effective method for sub-pixel disparity distribution learning, in which a sub-pixel level cost volume construction approach and an uncertainty-aware focal loss are developed to supervise the learning of disparity distribution.

\item Extensive experimental evaluations demonstrate the effectiveness of our approach. As shown in Fig. \ref{fig: intro} and Table \ref{table: avg}, our model outperforms the state-of-the-art LF depth estimation method\cite{chen2021attention} surpasses by \textbf{21.7\%} on BadPix 0.01, \textbf{28.5\%} on BadPix 0.03, \textbf{23.9\%} on BadPix 0.07, and \textbf{8.1\%} on MSE $\times$100, respectively.
\end{itemize}

The rest of this paper is organized as follows. Section II overviews the related work on LF depth map estimation. Section III provides the network components and loss functions of our method in detail. Section IV reports the experimental results on both synthetic and real LF datasets. Section V describes the failure cases and limitations of our method. Finally, the paper is concluded in Section VI.

\section{Related Work}
\label{sec:related}


\subsection{Light Field Visualization}
\begin{figure}[tb]
  \centering
   \includegraphics[width=\linewidth]{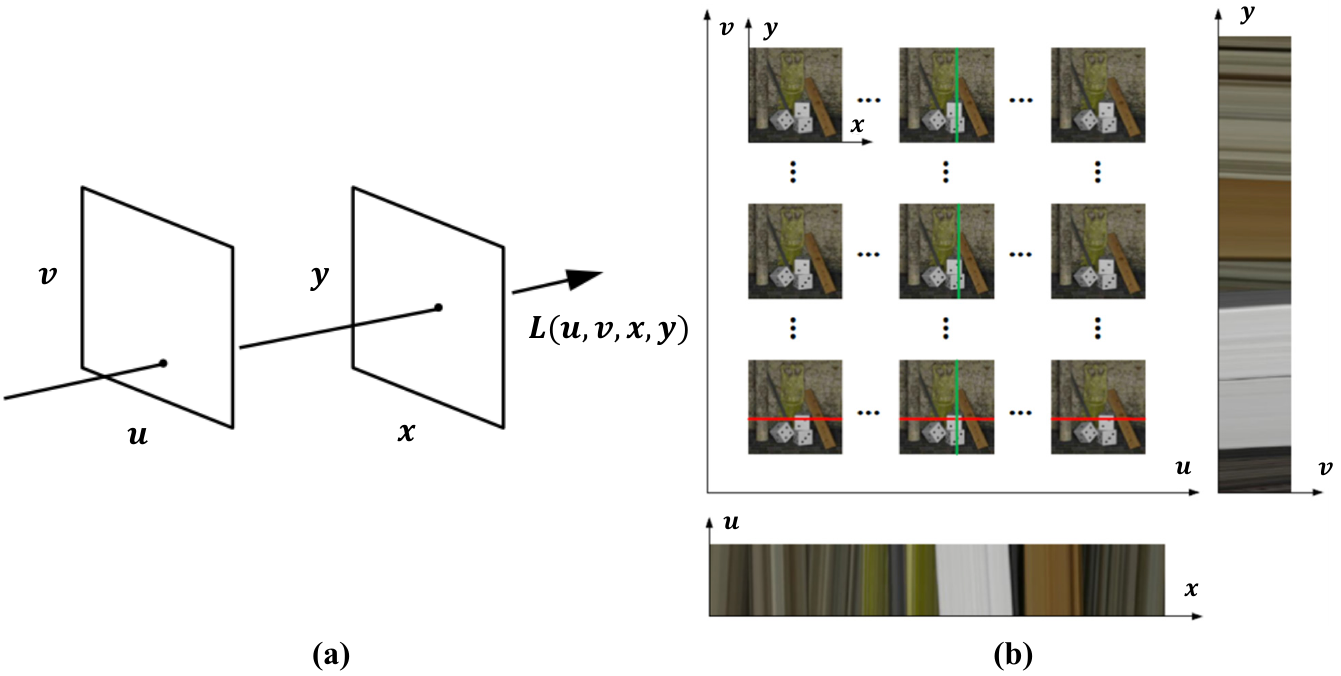}
   \caption{Light field visualization. (a) The two-plane LF model. (b) SAI and EPI representation of the LF image.}
   \label{fig: lf_vis}
\end{figure}

There is spatial and angular information in the LF image, which is often parametrically represented as $L(u,v,x,y)$ by a two-plane \cite{levoy1996light}. As shown in Fig.~\ref{fig: lf_vis} (a), where $(u,v)$ is the first plane, representing the angular resolution, and $(x,y)$ is the second plane, representing the spatial resolution. Based on the characteristics of the LF image, the LF can be represented as EPI or SAI as shown in Fig.~\ref{fig: lf_vis} (b).
We can obtain scene depth based on different representations of LF images. 

\subsection{Traditional Methods in LF Depth Estimation}

Traditional depth estimation algorithms can be divided into the following three categories based on the properties of LF images:  

\noindent \textbf{Multi-view stereo matching (MVS) based methods.} The LF image contains multi-view information from the scene, some works employ disparity cues by performing stereo matching from the LF image for depth estimation. Jeon \etal  \cite{jeon2015accurate} used phase shift theory to represent the sub-pixel translation between sub-aperture images. The central sub-aperture image was matched with other sub-aperture images to perform stereo matching. 

\noindent \textbf{Epipolar plane image (EPI) based methods.} Other methods calculate the slope of EPI for depth estimation. Wanner \etal  \cite{wanner2013variational} proposed a structure tensor to estimate the slope of lines in horizontal and vertical EPIs, and refined the initial results by global optimization. Zhang \etal  \cite{zhang2016robust} proposed a Spinning Parallelogram Operator (SPO) to compute the slope of straight lines in EPI, which is insensitive to occlusion, noise, and spatial blending. Chen \etal  \cite{chen2018accurate} performed superpixel regularization over partially occluded regions based on the characterization of EPI. Zhang \etal \cite{zhang2016light} exploited the special linear structure of an EPI and locally linear embedding (LLE).

\noindent \textbf{Defocus-based methods.} The defocused-based method performs depth estimation by calculating the color consistency of the views at different focal lengths. Tao \etal  \cite{tao2013depth} fused scattering and matching cues to obtain a local depth map using Markov random field for global optimization. Williem \etal  \cite{williem2016robust} employed the information entropy between different angles and adaptive scattering to improve robustness to occlusion and noise. An occlusion-aware vote cost (OAVC) method was presented by Han \etal \cite{han2021novel} to preserve depth map edges.
However, these methods rely on manually designed features and subsequent optimization, which are time-consuming with limited accuracy. In comparison, the proposed approach can fully utilize the power of deep networks to achieve higher accuracy and faster speed by learning the sub-pixel disparity distribution.

\subsection{Deep Learning-based Methods in LF Depth Estimation}
 
In recent years, deep learning has been gradually used for various tasks \cite{he2018learning, he2018spindle, li2022robust}, e.g., LF depth estimation \cite{heber2016convolutional, shin2018epinet, chen2021attention, tsai2020attention, wang2022occlusion}. Some learning-based methods estimate the slopes of the lines from the EPI. Luo \etal \cite{luo2017epi} used pairs of EPI image patches (horizontal and vertical) to train a CNN network. Li \etal \cite{li2020epi} designed a network to learn the relationship between the slopes of straight lines in horizontal and vertical EPIs. Leistner \etal \cite{leistner2019learning} introduced the idea of EPI-Shift to increase the receptive field of the network by virtually shifting the LF stack. These EPI-based methods only consider the features in the horizontal and vertical directions of angular views and require complex subsequent refinement.

Other methods design networks by directly exploring the correspondence among views in LF images. Shin \etal \cite{shin2018epinet} introduced a multi-stream input structure to concatenate views from four directions for depth estimation.
Guo \etal  \cite{guo2020accurate} proposed an occlusion-aware network that learns occlusion maps from features from multiple views, capable of accurately estimating the depth of sharp edges. 
Tsai \etal \cite{tsai2020attention} proposed an attention-based viewpoint selection network that generates attention maps to represent the importance of each viewpoint. Chen \etal \cite{chen2021attention} proposed an attention-based multi-level fusion network, using an intra-branch and inter-branch fusion strategy to perform a hierarchical fusion of effective features from different perspectives. Wang \etal \cite{wang2022occlusion} constructed an occlusion-aware cost volume based on dilated convolution \cite{yu2015multi}, which achieved a better trade-off between accuracy and speed.

The state-of-art methods \cite{tsai2020attention, chen2021attention, wang2022occlusion}  treat depth estimation as a regression problem supervised by a pixel-wise L1 loss and lack explicit supervision of disparity distribution. Different from previous methods, we directly learn sub-pixel disparity distribution for LF depth estimation, which can generate a more accurate depth map. To the best of our knowledge, we are the first to apply this methodology to LF depth estimation.

\subsection{Loss Functions in Depth Estimation}

The loss functions commonly used in supervised depth estimation methods include L1 Loss (MAE Loss), L2 Loss (MSE Loss), and Smooth L1 Loss. L2 Loss calculates the mean squared error between the predicted depth map and the actual depth label. The L1 Loss calculates the mean absolute error and has better robustness than L2 Loss to outliers. Smooth L1 Loss is a smoothed version of L1 Loss, which exploits the advantages of L1 and L2. Recently, Li \etal \cite{li2022binsformer} proposed the BinsFormer for the classification-regression-based monocular depth estimation, which can generate proper adaptive bins and ensure sufficient interaction between probability distribution and bin predictions. However, BinFormer's loss functions only consist of classification loss and regression loss, without considering depth distribution differences. Peng \etal \cite{peng2022rethinking} proposed a unified depth representation and a unified focal loss, which can be more uniform and reasonable to alleviate sample imbalance. However, a unified focal loss is based on a cross-entropy classification loss and lacks explicit supervision of disparity distribution. Our UAFL is inspired by Focal Loss \cite{lin2017focal, li2020generalized}. Different from previous studies, we introduced UAFL to address sample imbalance and adopted the dynamic weight adjustment strategy to make the model focus more on the hard points with significant differences in disparity distributions.

\begin{figure*}[tb]
  \centering
  \includegraphics[width=\linewidth]{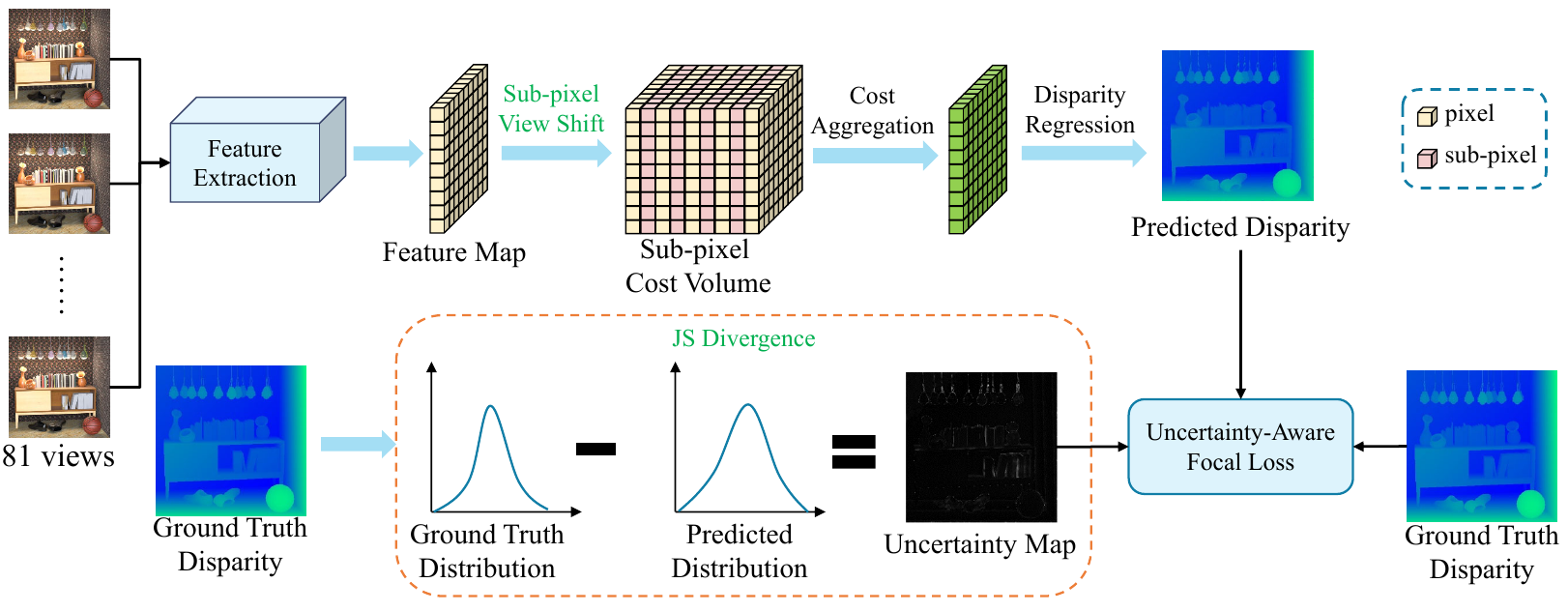}

  \caption{The architecture of our SubFocal network. A feature extractor is used to extract the features of each SAI and form the feature map. Then, a sub-pixel cost volume is constructed through a sub-pixel view shift based on bilinear interpolation. Next, we employ a cost aggregation module to fuse information on sub-pixel cost volume. In addition, a disparity regression module is adopted to produce the predicted disparity distribution and the disparity map. The uncertainty map is derived by calculating the JS divergence between the predicted disparity distribution and the ground truth. Finally, the proposed uncertainty-aware focal loss is employed to supervise disparity distribution and the disparity map.}
  \label{fig: network}
\end{figure*}

\section{Method}
\label{sec:method}

The architecture of our proposed approach is shown in Fig.~\ref{fig: network}. First, the features of each SAI are extracted using a shared feature extractor. Second, the sub-pixel view shift is performed to construct the sub-pixel cost volume. Third, the cost aggregation module is used to aggregate the cost volume information. The predicted disparity distribution and disparity map are then produced by attaching a disparity regression module, where the Jensen-Shannon (JS) divergence is employed to calculate the uncertainty map. Finally, the proposed uncertainty-aware focal loss is used to simultaneously supervise the disparity distribution and the disparity map. We will describe each module in detail below.

\subsection{Network}

\subsubsection{Feature Extraction}
First, two 3 $\times$ 3 convolutions (i.e., \textit{Conv2D\_1} and \textit{Conv2D\_2)} are employed to extract the initial feature with a channel of 4.
In order to extract multi-scale features of SAI, the widely used Spatial Pyramid Pooling (SPP) module  \cite{tsai2020attention,chen2021attention} is adopted as the feature extraction module in our network. Four average pooling operations at different scales are used to compress the features. The sizes of the average pooling blocks are 2 $\times$ 2, 4 $\times$ 4, 8 $\times$ 8, and 16 $\times$ 16, respectively. A 1 $\times$ 1 convolution layer is used for reducing the feature dimension for each scale. Bilinear interpolation is adopted to upsample these low-dimensional feature maps to the same size. The features of the SPP module contain hierarchical context information and are concatenated to form the feature map $F$. The hierarchical feature extraction strategy incorporates additional useful features from neighboring regions for challenging scenarios, such as texture-less and reflection areas.

\begin{figure}[tb]
  \centering
  \includegraphics[width=0.8\linewidth]{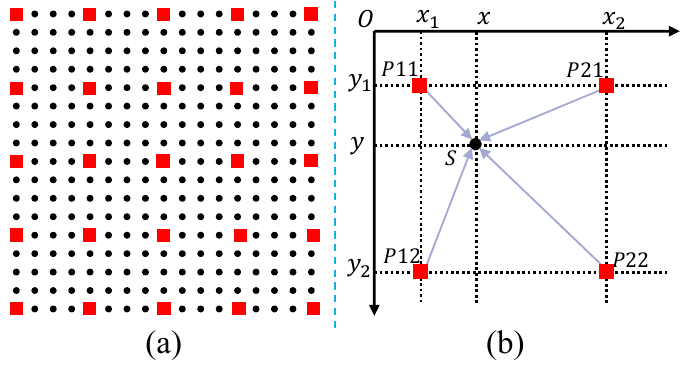}

  \caption{An illustration of constructing sub-pixel cost volume based on bilinear interpolation of the feature map level. Red points and black points stand for pixel level and sub-pixel level, respectively.}
  \label{fig: sub_shift}
\end{figure}

\subsubsection{Sub-pixel Cost Volume}
The LF can be represented using the two-plane model, i.e., $L(u,v,x,y)$, where $(x, y)$ stands for the spatial coordinate and $(u,v)$ for the angular coordinate. The LF disparity structure can be formulated as:

\begin{equation}
\label{eq: light}
\begin{aligned}
L(u_c,v_c,x,y)=L(u,v,x+(u-u_c)\cdot d(x,y), \\
y+(v-v_c)\cdot d(x,y)),
\end{aligned}
\end{equation}

\noindent where $d(x,y)$ denotes the disparity between the center view and its adjacent view at $(x, y)$.

Cost volume is constructed by shifting the feature according to Eq. \ref{eq: light} within a predefined disparity range (such as from -4 to 4). It can be transformed into the disparity distribution by subsequent cost aggregation and disparity regression modules. To obtain a finer disparity distribution, a sub-pixel level disparity sampling interval is required. Different from the previous method \cite{jeon2015accurate} using phase shift theorem to construct image level cost volume, we construct a sub-pixel feature level cost volume based on bilinear interpolation, which can save memory-consuming, as shown in Fig~\ref{fig: sub_shift}. Suppose the shape of the feature map $F$ is $H \times W \times C$. We want to shift the point $P11$ to point $S$, the shift distance is sub-pixel. The feature $F_S$ of the point $S$ can be calculated by bilinear interpolation:

\begin{equation}
\label{eq:bilinear}
\begin{aligned}
F_S & = \frac{(x_2-x)(y_2-y)} {(x_2-x_1)(y_2-y_1)} F_{P11} +\frac{(x-x_1)(y_2-y)} {(x_2-x_1)(y_2-y_1)} F_{P21} \\
& + \frac{(x_2-x)(y-y_1)} {(x_2-x_1)(y_2-y_1)} F_{P12} +\frac{(x-x_1)(y-y_1)} {(x_2-x_1)(y_2-y_1)} F_{P22}
\end{aligned}
\end{equation}

\noindent where the coordinates of $S$ are $(x,y)$. $P11$, $P12$, $P21$ and $P22$ are the four adjacent points, whose coordinates are $(x_1,y_1)$, $(x_1,y_2)$, $(x_2,y_1)$, and $(x_2,y_2)$, respectively.

Experimental results in  Sec. \ref{ablation: inter}  demonstrate that bilinear interpolation can achieve improved accuracy with higher efficiency. In our default setting, we adopt 17 disparity levels ranging from -4 to 4, where the sub-pixel interval is 0.5. After shifting the feature maps, we concatenate these feature maps into a 4D cost volume $D\times H\times W \times C$. 

It is worth noting that a smaller sampling interval can generate a finer disparity distribution but will increase computation time and slow down inference. Therefore, we conduct ablation experiments to investigate the trade-off between accuracy and speed with respect to the sampling interval, which is described in detail in  Sec. \ref{ablation: interval}.

\subsubsection{Cost Aggregation and Disparity Regression}
The shape of the sub-pixel cost volume is $D \times H \times W \times C$, and we employ 3D CNN to aggregate the sub-pixel cost volume. Following  \cite{tsai2020attention}, our cost aggregation consists of eight $3\times3\times3$ convolutional layers and two residual blocks. After passing through these 3D convolutional layers, we obtain the final cost volume $C_{f} \in  D \times H \times W$. We normalize $C_{f}$ by using the softmax operation along dimension $D$ to produce the probability of the disparity distribution $\hat{Y}_{dist}=\mathrm{softmax} (-C_f)$. Finally, the output disparity $\hat{d}$ can be calculated according to

\begin{equation}
\label{eq: softmin}
\begin{aligned}
\hat{d}=\sum_{d_k=D_{min}}^{D_{max}}d_k \times \mathrm{softmax} (-C_{d_k}),
\end{aligned}
\end{equation}

\noindent where $\hat{d}$ denotes the estimated center view disparity, $D_{min}$ and $D_{max}$ stand for the predefined minimum and maximum disparity values, respectively, $d_k$ is the disparity value between $D_{min}$ and $D_{max}$ according to the predefined disparity interval, and $C_{d_k}$ is the cost for the disparity value $d_k$.

\begin{figure}[tb]
  \centering
  \includegraphics[width=\linewidth]{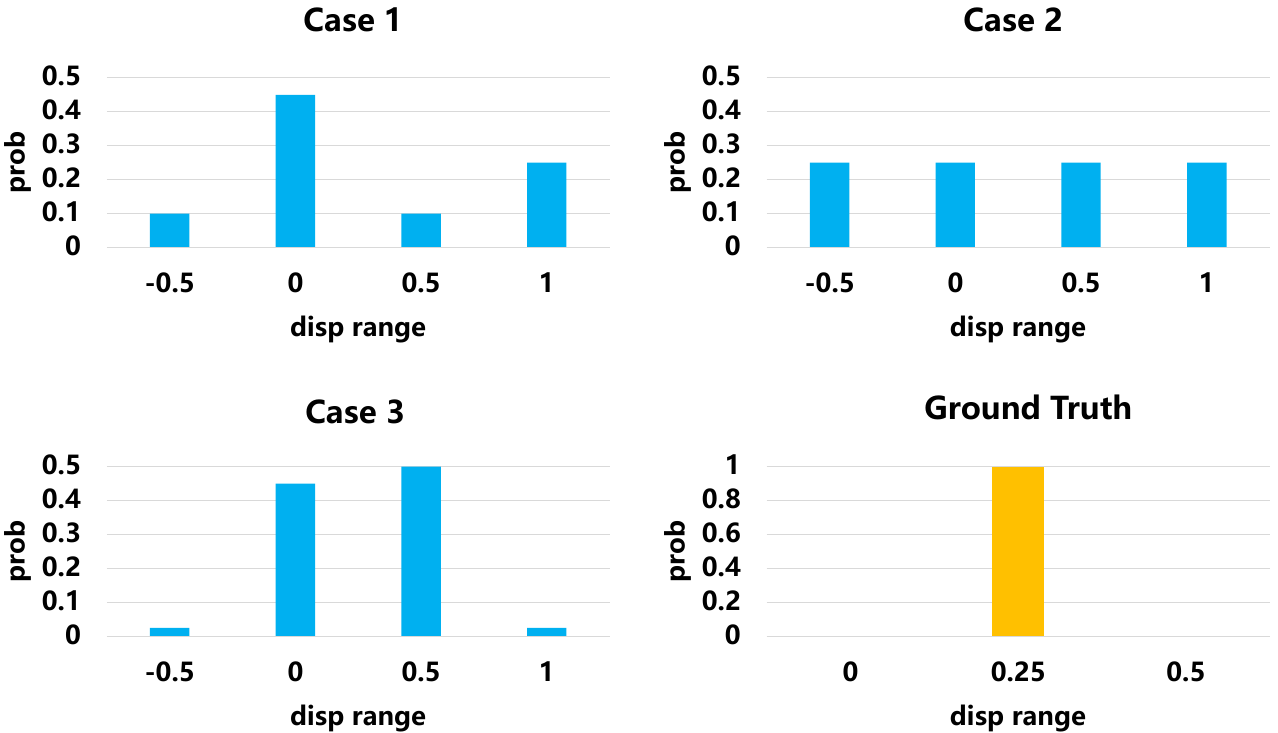}

  \caption{Example illustrating that L1 loss is not aware of the difference of disparity distribution. The bottom right sub-figure is the ground truth. Case 1, Case 2, and Case 3 are predicted disparity distributions.}
  \label{fig:dist}
\end{figure}

\subsection{Uncertainty-aware Focal Loss}

\begin{figure}[tb]
  \centering
  \includegraphics[width=\linewidth]{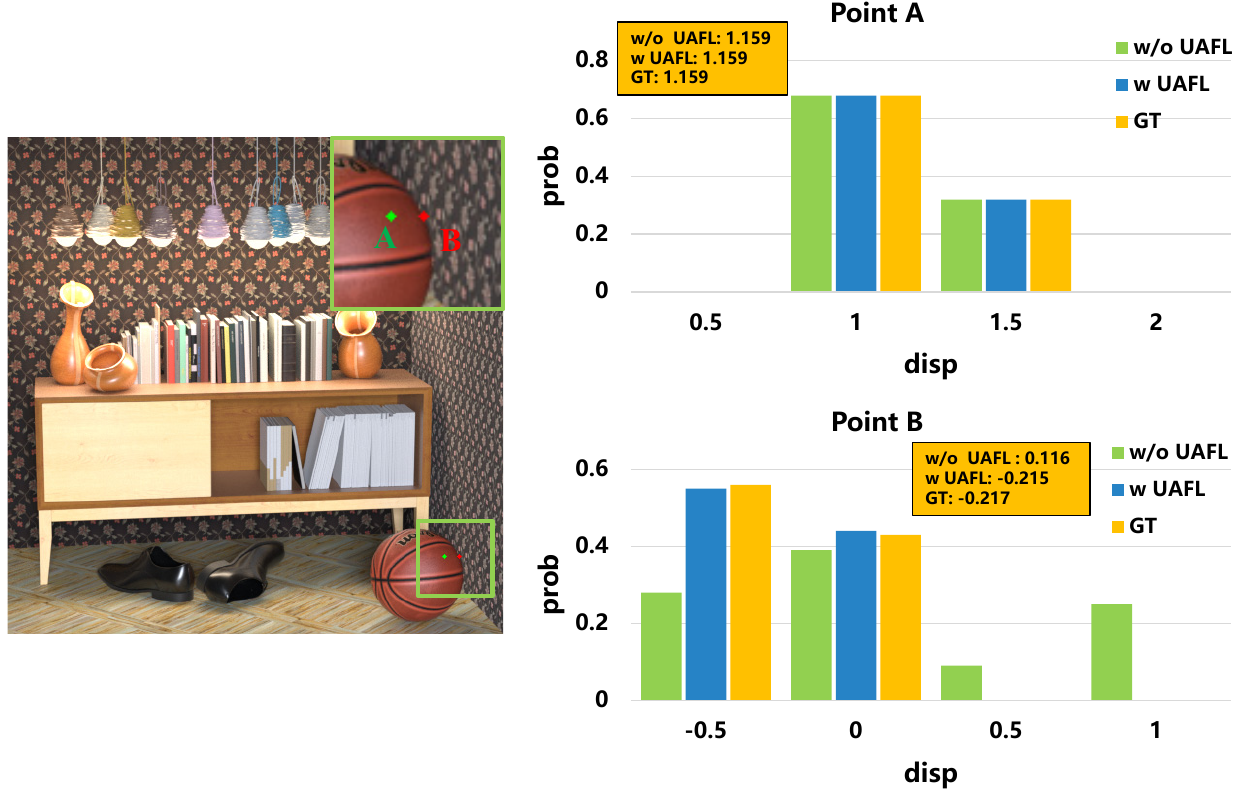}

    \caption{Visualization of disparity distribution. We compare the disparity distribution of points on $Sideboard$ with/without using UAFL. The disparity distribution with UAFL is closer to the ground truth in the occlusion region.}
  \label{fig: vis_dist}
\end{figure}

Previous methods \cite{shin2018epinet, tsai2020attention,chen2021attention, wang2022occlusion} consider depth estimation as a simple regression process, and treat each point equally via a pixel-wise L1 loss. 
Fig.~\ref{fig:dist} shows the example of disparity distributions. From the example, we can see that unlimited disparity distributions may produce the same expectation of disparity, e.g. 0.25, however, only the one close to the ground truth is rational, i.e. Case 3.
We claim that a desirable model needs to learn the reasonable disparity distributions (e.g., Dirac delta distribution \cite{li2020generalized}) for depth estimation. It is challenging to use the simple L1 loss to supervise the disparity distribution, especially for those difficult spatial areas. Here, we visualize the disparity distributions of our method in Fig.~\ref{fig: vis_dist}. In simple and non-occlusion regions (i.e., Point $A$), a reasonable disparity distribution and accurate disparity can be obtained by directly using L1 loss for training. In challenging occlusion regions (i.e., Point $B$), however, it struggles to obtain a reasonable disparity distribution and accurate disparity directly from L1 loss. In addition, the image generally contains a large number of non-occlusion regions and a small number of occlusion regions. It highlights an imbalance problem in learning disparity distribution and disparity.

To address this issue, we design an uncertainty-aware focal loss to supervise the disparity distribution. Similar to Focal Loss \cite{lin2017focal, li2020generalized}, we adopt the dynamic weight adjustment strategy to make the model focus more on the hard points with significant differences in disparity distributions. Fig.~\ref{fig: vis_dist} compares the disparity distribution with and without using UAFL. By using UAFL, our model can predict a reasonable disparity distribution and accurate disparity in challenge regions. The process of UAFL is detailed below.

In order to quantitatively evaluate the difficulty level of different regions, we adopt the JS divergence to measure the difference between the predicted and the ground-truth disparity distributions as presented in Eq. \ref{eq: JS}.

\begin{equation}
\label{eq: JS}
\begin{aligned}
U = JS(Y_{dist}||\hat{Y}_{dist}),
\end{aligned}
\end{equation}
 
 \noindent where $U$ is the uncertainty map, $\hat{Y}_{dist}$ represents the predicted disparity distribution, and $Y_{dist}$ is the ground-truth disparity distribution. We follow \cite{li2020generalized} to consider the ground-truth disparity distribution as a Dirac delta distribution and discretize it to the two adjacent sampling points $d_l$ and $d_r$, according to:
 
\begin{equation}
\label{eq: disc}
\begin{aligned}
Y_{dist} = p_{d_l}d_l +p_{d_r}d_r,
\end{aligned}
\end{equation}

\noindent where $d_l$ and $d_r$ are the left and right adjacent points respectively, and $d$ denotes the ground-truth disparity. $p_{d_l}=\frac{d_r-d}{d_r-d_l}$ and $p_{d_r}=\frac{d-d_l}{d_r-d_l}$ represent their corresponding probability values. 
We employ discretization for two reasons. Firstly, it approximates the actual disparity distribution (Dirac delta) more closely. Secondly, it ensures that the expected value of the distribution remains constant before and after discretization.

Finally, we design an uncertainty-aware focal loss to supervise both disparity distribution and disparity map: 

\begin{equation}
\label{eq: uafl}
\begin{aligned}
UAFL=\left \| U \right \| ^{\beta } \odot \left \| d - \hat{d} \right \| _1,
\end{aligned}
\end{equation}

\noindent where $U$ is the uncertainty map, $\hat{d}$ is the predicted disparity, and $d$ is the ground-truth disparity. $\odot$ stands for the element-wise multiplication operation, and $\beta$ is the coefficient factor that controls the ratio of dynamic weight assignment. When $\beta=0$, UAFL degenerates to the standard L1 loss. In ablation experiments of Sec. \ref{ablation: beta}, we analyze the effect of different $\beta$ values.

\begin{table*}[ht]
\centering
\caption{Quantitative comparison results with the top-ranked methods on HCI 4D benchmark, containing the SPO \cite{zhang2016robust}, CAE \cite{park2017robust},  Epinet-fcn-m \cite{shin2018epinet}, EPI\_ORM \cite{li2020epi}, LFattNet \cite{tsai2020attention}, AttMLFNet \cite{chen2021attention}, FastLFNet \cite{huang2021fast}, DistgDisp \cite{wang2022disentangling}, OACC-Net \cite{wang2022occlusion} and SubFocal(Ours). The best results are in bold faces and the second-best results are underlined.}
\renewcommand\arraystretch{1.1}
\resizebox{2\columnwidth}{!}{
\begin{tabular}{l|cccc|cccc|cccc|cccc}
\toprule
 & \multicolumn{4}{c|}{Backgammon} & \multicolumn{4}{c|}{Dots} & \multicolumn{4}{c|}{Pyramids} & \multicolumn{4}{c}{Stripes} \\ \cline{2-17} 
 & 0.07 & 0.03 & 0.01 & MSE & 0.07 & 0.03 & 0.01 & MSE & 0.07 & 0.03 & 0.01 & MSE & 0.07 & 0.03 & 0.01 & MSE \\ \midrule
SPO \cite{zhang2016robust} & 3.781 & 8.639 & 49.94 & 4.587 & 16.27 & 35.06 & 58.07 & 5.238 & 0.861 & 6.263 & 79.20 & 0.043 & 14.99 & 15.46 & 21.87 & 6.955 \\
CAE \cite{park2017robust} & 3.924 & 4.313 & 17.32 & 6.074 & 12.40 & 42.50 & 83.70 & 5.082 & 1.681 & 7.162 & 27.54 & 0.048 & 7.872 & 16.90 & 39.95 & 3.556 \\
PS\_RF \cite{jeon2018depth} & 7.142 & 13.94 & 74.66 & 6.892 & 7.975 & 17.54 & 78.80 & 8.338 & \textbf{0.107} & 6.235 & 83.23 & 0.043 & 2.964 & 5.790 & 41.65 & 1.382 \\
OBER-cross-ANP \cite{schilling2018trust} & 3.413 & 4.956 & 13.66 & 4.700 & 0.974 & 37.66 & 73.13 & 1.757 & 0.364 & 1.130 & 8.171 & 0.008 & 3.065 & 9.352 & 44.72 & 1.435 \\
OAVC \cite{han2021novel} & {\ul 3.121} & 5.117 & 49.05 & 3.835 & 69.11 & 75.38 & 92.33 & 16.58 & 0.831 & 9.027 & 33.66 & 0.040 & 2.903 & 19.88 & 28.14 & 1.316 \\
Epinet-fcn-m \cite{shin2018epinet} & 3.501 & 5.563 & 19.43 & 3.705 & 2.490 & 9.117 & 35.61 & 1.475 & 0.159 & 0.874 & 11.42 & 0.007 & {\ul 2.457} & {\ul 2.711} & 11.77 & 0.932 \\
EPI\_ORM \cite{li2020epi} & 3.988 & 7.238 & 34.32 & \textbf{3.411} & 36.10 & 47.93 & 65.71 & 14.48 & 0.324 & 1.301 & 19.06 & 0.016 & 6.871 & 13.94 & 55.14 & 1.744 \\
LFattNet \cite{tsai2020attention} & 3.126 & {\ul 3.984} & {\ul 11.58} & {\ul 3.648} & 1.432 & 3.012 & 15.06 & 1.425 & 0.195 & 0.489 & 2.063 & {\ul 0.004} & 2.933 & 5.417 & 18.21 & 0.892 \\
AttMLFNet \cite{chen2021attention} & 3.228 & 4.625 & 13.73 & 3.863 & 1.606 & 2.021 & {\ul 10.61} & \textbf{1.035} & 0.174 & {\ul 0.429} & \textbf{ 1.767} & \textbf{0.003} & 2.932 & 4.743 & 15.44 & \textbf{0.814} \\
FastLFNet \cite{huang2021fast} & 5.138 & 11.41 & 39.84 & 3.986 & 21.17 & 41.11 & 68.15 & 3.407 & 0.620 & 2.193 & 22.19 & 0.018 & 9.442 & 32.60 & 63.04 & 0.984 \\
DistgDisp \cite{wang2022disentangling} & 5.824 & 10.54 & 26.17 & 4.712 & 1.826 & 4.464 & 25.37 & 1.367 & {\ul 0.108} & 0.539 & 4.953 & {\ul 0.004} & 3.913 & 6.885 & 19.25 & 0.917 \\
OACC-Net \cite{wang2022occlusion} & 3.931 & 6.640 & 21.61 & 3.938 & 1.510 & 3.040 & 21.02 & 1.418 & 0.157 & 0.536 & 3.852 & {\ul 0.004} & 2.920 & 4.644 & 15.24 & 0.845 \\
SubFocal(Ours) & 3.194 & 4.281 & 12.47 & 3.667 & \textbf{0.899} & {\ul 1.524} & 15.51 & 1.301 & 0.220 & \textbf{0.411} & {\ul 1.867} & 0.005 & 2.464 & 3.568 & {\ul 9.386} & {\ul 0.821} \\
SubFocal-L(Ours) & \textbf{3.079} & \textbf{ 3.651} & \textbf{7.821} & 3.868 & {\ul 0.935} & \textbf{1.133} & \textbf{8.535} & {\ul 1.279} & 0.253 & 0.543 & 2.017 & 0.005 & \textbf{2.122} & \textbf{2.219} & \textbf{3.992} & 0.874 \\ 
\bottomrule
\toprule
 & \multicolumn{4}{c|}{Boxes} & \multicolumn{4}{c|}{Cotton} & \multicolumn{4}{c|}{Dino} & \multicolumn{4}{c}{Sideboard} \\ \cline{2-17} 
 & 0.07 & 0.03 & 0.01 & MSE & 0.07 & 0.03 & 0.01 & MSE & 0.07 & 0.03 & 0.01 & MSE & 0.07 & 0.03 & 0.01 & MSE \\ 
\midrule
SPO \cite{zhang2016robust} & 15.89 & 29.52 & 73.23 & 9.107 & 2.594 & 13.71 & 69.05 & 1.313 & 2.184 & 16.36 & 69.87 & 0.310 & 9.297 & 28.81 & 73.36 & 1.024 \\
CAE \cite{park2017robust} & 17.89 & 40.40 & 72.69 & 8.424 & 3.369 & 15.50 & 59.22 & 1.506 & 4.968 & 21.30 & 61.06 & 0.382 & 9.845 & 26.85 & 56.92 & 0.876 \\
PS\_RF \cite{jeon2018depth} & 18.95 & 35.23 & 76.39 & 9.043 & 2.425 & 14.98 & 70.41 & 1.161 & 4.379 & 16.44 & 75.97 & 0.751 & 11.75 & 36.28 & 79.98 & 1.945 \\
OBER-cross-ANP \cite{schilling2018trust} & 10.76 & 17.92 & 44.96 & 4.750 & 1.108 & 7.722 & 36.79 & 0.555 & 2.070 & 6.161 & 22.76 & 0.336 & 5.671 & 12.48 & 32.79 & 0.941 \\
OAVC \cite{han2021novel} & 16.14 & 33.68 & 71.91 & 6.988 & 2.550 & 20.79 & 61.35 & 0.598 & 3.936 & 19.03 & 61.82 & 0.267 & 12.42 & 37.83 & 73.85 & 1.047 \\
Epinet-fcn-m \cite{shin2018epinet} & 12.34 & 18.11 & 46.09 & 5.968 & 0.447 & 2.076 & 25.72 & 0.197 & 1.207 & 3.105 & 19.39 & 0.157 & 4.462 & 10.86 & 36.49 & 0.798 \\
EPI\_ORM \cite{li2020epi} & 13.37 & 25.33 & 59.68 & 4.189 & 0.856 & 5.564 & 42.94 & 0.287 & 2.814 & 8.993 & 41.04 & 0.336 & 5.583 & 14.61 & 52.59 & 0.778 \\
LFattNet \cite{tsai2020attention} & 11.04 & 18.97 & 37.04 & 3.996 & 0.272 & 0.697 & 3.644 & 0.209 & 0.848 & 2.340 & 12.22 & 0.093 & 2.870 & 7.243 & 20.73 &  0.530 \\
AttMLFNet \cite{chen2021attention} & 11.14 & 18.65 & 37.66 & 3.842 & \textbf{0.195} & \textbf{0.374} & \textbf{1.522} & \textbf{0.059} & \textbf{0.440} & \textbf{ 1.193} & \textbf{4.559} & \textbf{0.045} & {\ul 2.691} & 6.951 & 21.56 & \textbf{0.398} \\
FastLFNet \cite{huang2021fast} & 18.70 & 37.45 & 71.82 & 4.395 & 0.714 & 6.785 & 49.34 & 0.322 & 2.407 & 13.27 & 56.24 & 0.189 & 7.032 & 21.62 & 61.96 & 0.747 \\
DistgDisp \cite{wang2022disentangling} & 13.31 & 21.13 & 41.62 & 3.325 & 0.489 & 1.478 & 7.594 & 0.184 & 1.414 & 4.018 & 20.46 & 0.099 & 4.051 & 9.575 & 28.28 & 0.713 \\
OACC-Net \cite{wang2022occlusion} & 10.70 & 18.16 & 43.48 & {\ul 2.892} & 0.312 & 0.829 & 10.45 & {\ul 0.162} & 0.967 & 2.874 & 22.11 & {\ul 0.083} & 3.350 & 8.065 & 28.64 & 0.542 \\
SubFocal(Ours) & {\ul 8.536} & {\ul 16.44} & {\ul 32.03} & 2.993 &0.257 & 0.611 & 3.337 & 0.188 & 0.777 & 2.052 & 10.23 & 0.141 & \textbf{2.360} & \textbf{6.113} & {\ul 18.95} & {\ul 0.404} \\ 
SubFocal-L(Ours) & \textbf{7.266} & \textbf{11.41} & \textbf{29.61} & \textbf{2.417} &{\ul 0.252} & {\ul 0.501} & {\ul 3.072} & 0.243 & {\ul 0.684} & {\ul 1.735} & {\ul 9.745} & 0.101 & 2.694 & {\ul 6.246} & \textbf{18.26} & 0.441 \\ 
\bottomrule
\toprule
 & \multicolumn{4}{c|}{Bedroom} & \multicolumn{4}{c|}{Bicycle} & \multicolumn{4}{c|}{Herbs} & \multicolumn{4}{c}{Origami} \\ \cline{2-17} 
 & 0.07 & 0.03 & 0.01 & MSE & 0.07 & 0.03 & 0.01 & MSE & 0.07 & 0.03 & 0.01 & MSE & 0.07 & 0.03 & 0.01 & MSE \\ 
\midrule
SPO \cite{zhang2016robust} & 4.864 & 23.53 & 72.37 & 0.209 & 10.91 & 26.90 & 71.13 & 5.570 & 8.260 & 30.62 & 86.62 & 11.23 & 11.69 & 32.71 & 75.58 & 2.032 \\
CAE \cite{park2017robust} & 5.788 & 25.36 & 68.59 & 0.234 & 11.22 & 23.62 & 59.64 & 5.135 & 9.550 & 23.16 & 59.24 & 11.67 & 10.03 & 28.35 & 64.16 & 1.778 \\
PS\_RF \cite{jeon2018depth} & 6.015 & 22.45 & 80.68 & 0.288 & 17.17 & 32.32 & 79.80 & 7.926 & 10.48 & 21.90 & 66.47 & 15.25 & 13.57 & 36.45 & 80.32 & 2.393 \\
OBER-cross-ANP \cite{schilling2018trust} & 3.329 & 9.558 & 28.91 & 0.185 & 8.683 & 16.17 & 37.83 & 4.314 & 7.120 & 14.06 & 36.83 & 10.44 & 8.665 & 20.03 & 42.16 & 1.493 \\
OAVC \cite{han2021novel} & 4.915 & 19.09 & 64.76 & 0.212 & 12.22 & 25.46 & 64.74 & 4.886 & 8.733 & 29.65 & 74.76 & 10.36 & 12.56 & 30.59 & 69.35 & 1.478 \\
Epinet-fcn-m \cite{shin2018epinet} & 2.299 & 6.345 & 31.82 & 0.204 & 9.614 & 16.83 & 42.83 & 4.603 & 10.96 & 25.85 & 59.93 & 9.491 & 5.807 & 13.00 & 42.21 & 1.478 \\
EPI\_ORM \cite{li2020epi} & 5.492 & 14.66 & 51.02 & 0.298 & 11.12 & 21.20 & 51.22 & 3.489 & 8.515 & 24.60 & 68.79 & \textbf{4.468} & 8.661 & 22.95 & 56.57 & 1.826 \\
LFattNet \cite{tsai2020attention} & 2.792 & 5.318 &  13.33 & 0.366 & 9.511 & 15.99 & 31.35 & 3.350 & 5.219 & 9.473 & 19.27 & 6.605 & 4.824 & 8.925 & 22.19 & 1.733 \\
AttMLFNet \cite{chen2021attention} & {\ul 2.074} &  5.272 & 16.18 & 0.129 & 8.837 & 16.06 & 32.71 & 3.082 & 5.426 & 9.468 & 18.84 & 6.374 & 4.406 & 9.032 &  22.45 & 0.991 \\
FastLFNet \cite{huang2021fast} & 4.903 & 15.92 & 52.88 & 0.202 & 15.38 & 28.45 & 59.24 & 4.715 & 10.72 & 23.39 & 59.98 & 8.285 & 12.64 & 33.65 & 72.36 & 2.228 \\
DistgDisp \cite{wang2022disentangling} & 2.349 & 5.925 & 17.66 & \textbf{0.111} & 9.856 & 17.58 & 35.72 & 3.419 & 6.846 & 12.44 & 24.44 & 6.846 & 4.270 & 9.816 & 28.42 & 1.053 \\
OACC-Net \cite{wang2022occlusion} & 2.308 & 5.707 & 21.97 & 0.148 & 8.078 & 14.40 & 32.74 & 2.907 & 6.515 & 46.78 & 86.41 & 6.561 & 4.065 & 9.717 & 32.25 & \textbf{0.878} \\
SubFocal(Ours) & 2.234 & {\ul 4.364} & {\ul 11.81} & 0.141 & {\ul 7.277} & {\ul 12.75} & {\ul 28.39} & \textbf{2.670} & {\ul 4.297} & {\ul 8.190} & {\ul 17.36} & 6.126 & {\ul 2.961} & {\ul 6.917} & 19.33 & 0.956 \\
SubFocal-L(Ours) & \textbf{1.882} & \textbf{3.669} & \textbf{10.34} & {\ul 0.125} & \textbf{6.829} & \textbf{11.64} & \textbf{25.66} & {\ul 2.689} & \textbf{3.998} & \textbf{7.238} & \textbf{16.65} & {\ul 6.041} & \textbf{2.823} & \textbf{6.388} & \textbf{18.43} & {\ul 0.883} 
\\ 
\bottomrule
\end{tabular}

}

\label{table: quantitative}
\end{table*}

\begin{table*}[!tb]
\centering
\caption{The average BadPix 0.07, BadPix 0.03, BadPix 0.01, and MSE $\times$100 on the HCI 4D LF
benchmark. The best results are in bold faces and the second-best results are underlined.}
\renewcommand\arraystretch{1.2}
\resizebox{2\columnwidth}{!}{
\begin{tabular}{l|ccccccc}
\toprule
Average & SPO \cite{zhang2016robust}  & CAE \cite{park2017robust} & PS\_RF \cite{jeon2018depth}  & OBER-cross-ANP \cite{schilling2018trust} & OAVC \cite{han2021novel} & Epinet-fcn-m \cite{shin2018epinet} & EPI\_ORM \cite{li2020epi} \\ 
\midrule
BP 0.07 & 8.466 & 8.211 & 8.578 & 4.594 & 12.45 & 4.646 & 8.642 \\
BP 0.03 & 22.30 & 22.95 & 21.63 & 13.10 & 27.13 & 9.537 & 17.36 \\
BP 0.01 & 66.70 & 55.84 & 74.03 & 35.23 & 62.14 & 31.90 & 49.84 \\
MSE $\times$100 & 3.968 & 3.730 & 4.617 & 2.584 & 3.968 & 2.418 & 2.944 \\
\bottomrule
\toprule
Average & LFattNet \cite{tsai2020attention} & AttMLFNet \cite{chen2021attention} & FastLFNet \cite{huang2021fast} & DistgDisp \cite{wang2022disentangling} & OACC-Net \cite{wang2022occlusion} & SubFocal(Ours) & SubFocal-L(Ours) \\ 
\midrule
BP 0.07 & 3.756 & 3.596 & 9.071 & 4.521 & 3.734 & {\ul 2.956} & \textbf{2.735} \\
BP 0.03 & 6.823 & 6.568 & 22.32 & 8.699 & 10.12 & {\ul 5.602} & \textbf{4.697}  \\
BP 0.01 & 17.23 & 16.42 & 56.45 & 23.33 & 28.32 & {\ul 15.06} & \textbf{12.85}  \\
MSE $\times$100 & 1.904 & 1.720 & 2.456 & 1.896 & 1.698 & {\ul 1.618} & \textbf{1.581}  \\
\bottomrule

\end{tabular}
}
\label{table: avg}
\end{table*}

\section{Experiments}
\label{sec:exp}
In this section, we first describe the dataset, evaluation metrics, and implementation details, and then compare our method with state-of-the-art methods. Finally, we present ablation experiments to analyze the effectiveness of the proposed method.

\begin{table*}[tb]
\centering
\caption{Quantitative comparison results with state-of-the-art methods on the scenes of UrbanLF-Syn dataset\cite{sheng2022urbanlf} in terms of BadPix 0.07 and Runtime. Runtime is the average result of running each model 50 times with an input LF image of 9$\times$9$\times$256$\times$256. The best results are shown in boldface.
}
\renewcommand\arraystretch{1.0}
\resizebox{2.0 \columnwidth}{!}{
\begin{tabular}{l|c|cccccccc|cc}
\toprule
Method & Params. & Img11 & Img27 & Img34 & Img50 & Img54 & Img68 & Img69 & Img70 & BP 0.07 & Runtime \\ 
\midrule 
LFattNet & 5.06M & 10.179 & 23.648 & 22.637 & 9.668 & 13.533 & 9.819 & 12.075 & 7.467 & 13.629  & 0.743s \\
DistgDisp & 4.89M & \textbf{6.825} & \textbf{20.240} & 24.522 & 8.304 & 13.830 & 9.367 & 10.118 & 7.691 & 12.612 & 0.444s \\
OACC-Net & 5.02M & 6.845 & 24.058 & 27.792 & 10.390 & 13.922 & 11.799 & 13.234 & 8.750 & 14.599 & \textbf{0.439s} \\
Ours & 5.06M & 9.194 & 22.223 & \textbf{22.619} & \textbf{6.008} & \textbf{11.476} & \textbf{8.385} & \textbf{9.357} & \textbf{6.542} & \textbf{11.976} & 1.839s \\
\bottomrule
\end{tabular}}
\label{table: quantitative_urban}
\end{table*}

\begin{figure}[tb]
  \centering
  \includegraphics[width=\linewidth]{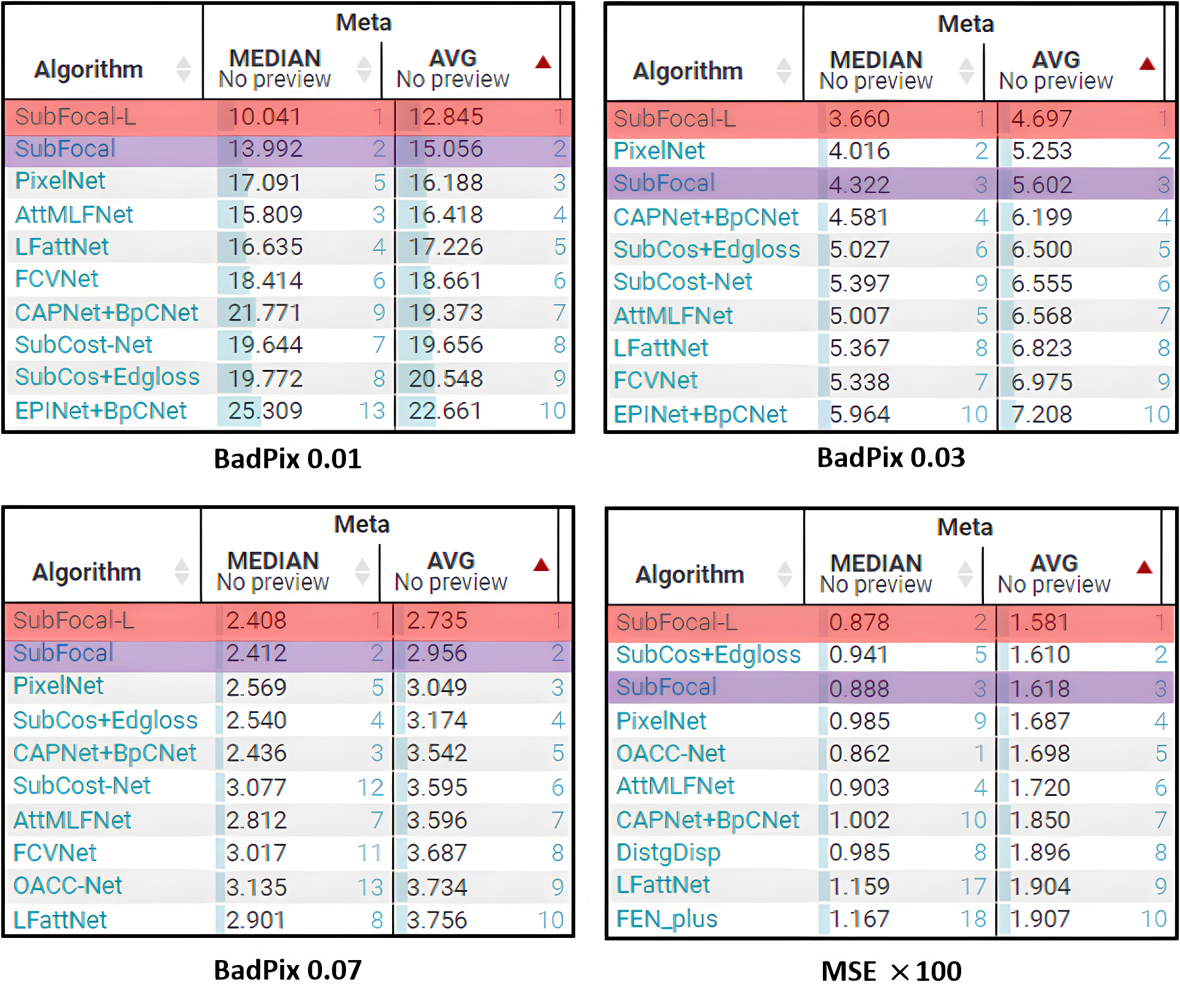}
  \caption{The screenshot of the rankings on the 4D LF benchmark [https://lightfield-analysis.uni-konstanz.de/]. The enhanced version (with a sampling rate of 0.1) of our method, called SubFocal-L, ranks first place among all the 105 submissions in terms of all four accuracy metrics, while the standard version  (with a sampling rate of 0.5) of our method (i.e., SubFocal) also achieves top-performance.}
  \label{fig: screenshot}
\end{figure}

\subsection{Datasets, Evaluation Metrics, and Implementation Details}
The 4D LF Benchmark (\textbf{HCI 4D})  \cite{honauer2016dataset} is the current mainstream benchmark dataset for the evaluation of LF disparity estimation methods. This dataset is rendered by Blender and provides 28 densely arranged 4D LFs, including 16 scenes in the 
\textit{Additional} category, 4 scenes in the \textit{Training} category, 4 scenes in the \textit{Test} category, and 4 scenes in the \textit{Stratified} category. Scenes in the \textit{Training}, \textit{Additional}, and \textit{Stratified} have ground-truth disparity maps while scenes in the \textit{Test} category do not provide ground-truth disparity annotations. The spatial resolution of each scene is 512$\times$512, and the angular resolution is 9$\times$9. \textbf{UrbanLF-Syn} dataset\cite{sheng2022urbanlf} contains 230 synthetic LF samples, with 170 training, 30 validation, and 30 test samples for LFNAT LF Depth Estimation Challenge at the CVPR 2023 Workshop. Each sample consists of 81 SAIs with a spatial resolution of 480$\times$640 and an angular resolution of 9$\times$9.

For the quantitative evaluation, we adopt the standard metrics, including MSE $\times$100 and BadPix($\epsilon $). Specifically, MSE $\times$100 represents the mean square errors of all pixels at a given mask multiplied by 100. BadPix($\epsilon $) denotes the percentage of pixels whose absolute value between the actual label at a given mask and the algorithm's predicted result surpasses a threshold $\epsilon $, which is usually chosen as 0.01, 0.03, and 0.07.


We follow the previous methods \cite{shin2018epinet,tsai2020attention,chen2021attention,wang2022occlusion} and use 16 scenes from \textit{Additional} as the training set, 8 scenes from \textit{Stratified} and \textit{Training} as the validation set, and 4 scenes from \textit{Test} as the test set. We employ LFattNet \cite{tsai2020attention} as the baseline model, change disparity interval, and loss functions. During training, LF images were randomly cropped into 32$\times$32 grayscale patches. The same data augmentation strategy \cite{shin2018epinet,tsai2020attention} was used to improve the model performance. We set the batchsize to 32 and used the Adam \cite{kingma2014adam} optimizer. The disparity range was set to [-4, 4], and the sampling interval was set to 0.5 by default (i.e., SubFocal). To speed up the model training, a two-step strategy was adopted. In the first stage, the model was trained for 50 epochs using L1 Loss, and the learning rate was set to 1$\times$10$^{-3}$. In the second stage, the L1 loss was replaced by our UAFL. The model was finetuned for 10 epochs, and the learning rate was set to 1$\times$10$^{-4}$. Our network is implemented using TensorFlow \cite{abadi2016tensorflow}. The training process took roughly 65 hours on an NVIDIA RTX 3090 GPU. We also introduced an enhanced model called SubFocal-L by setting the sampling interval to 0.1. SubFocal-L was trained for 165 hours on an NVIDIA A40 GPU, which was approximately equivalent to 330 hours on an NVIDIA RTX 3090 GPU. As for the Urban-Syn dataset, we follow the official splitting with 170 scenes for training and randomly select 8 scenes for testing. The disparity range was set to [-0.5, 1.6], and the sampling interval was set to 0.1. It took about 158 hours on an NVIDIA RTX 3090 GPU to train the model.

\begin{figure*}[!ht]
  \centering
  \includegraphics[width=0.95\linewidth]{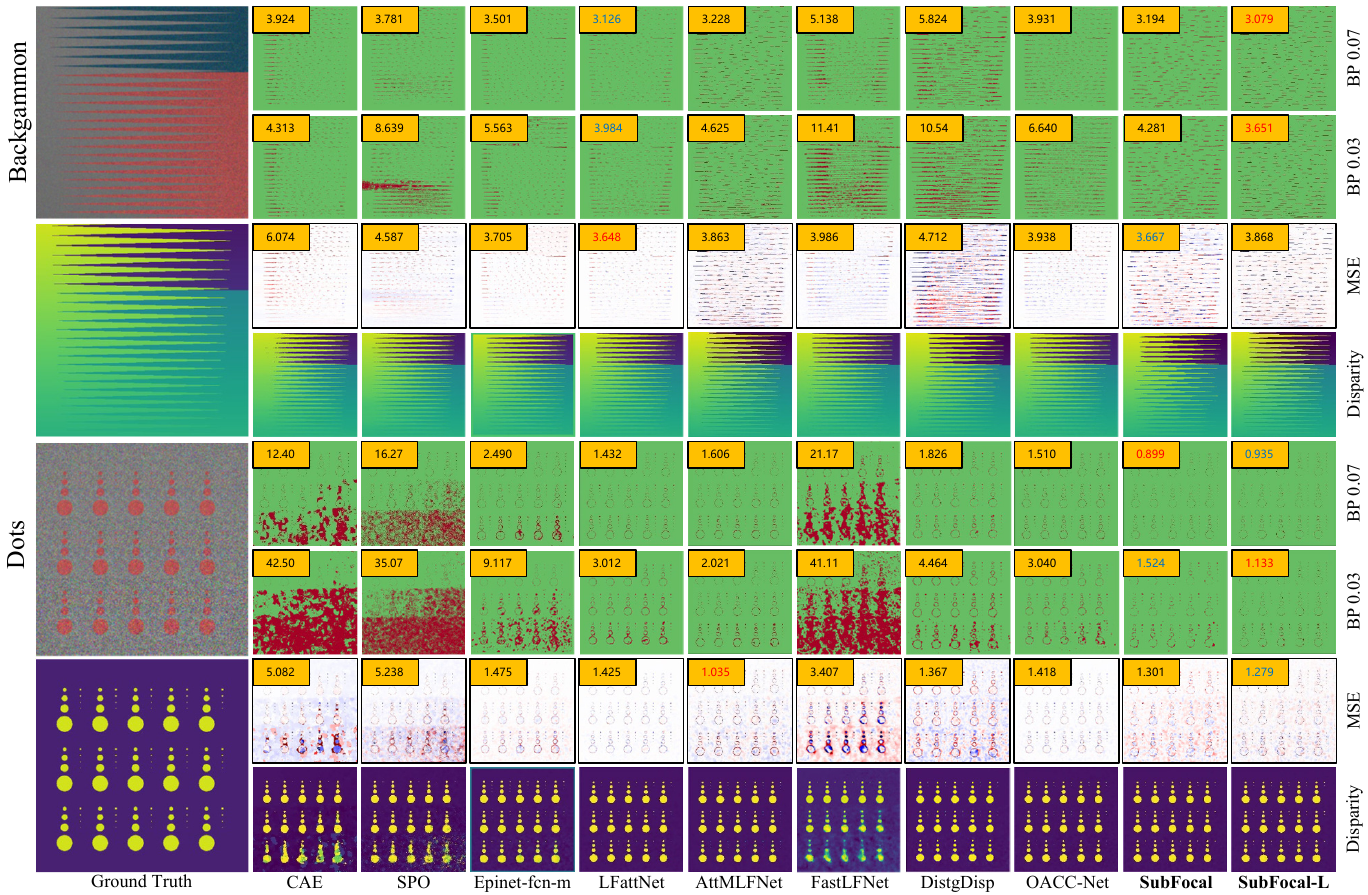}
  \includegraphics[width=0.95\linewidth]{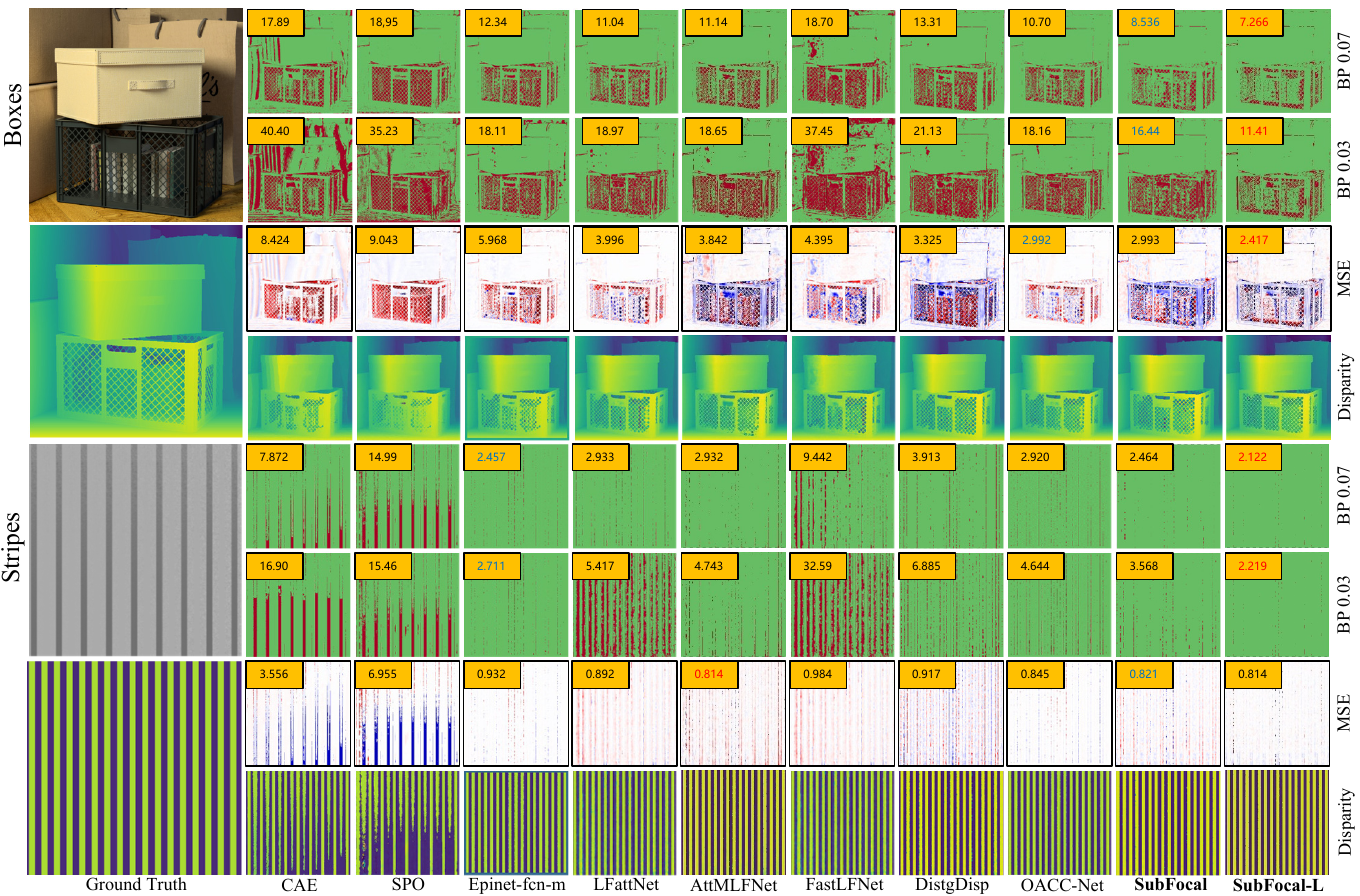}
  
    \caption{Visual comparisons of disparity and error maps on validation scenes \textit{Backgammon}, \textit{Dots}, \textit{Boxes}, and \textit{Stripes} \cite{rerabek2016new}. Corresponding quantitative scores (BadPix 0.07, BadPix 0.03, and MSE $\times$100) are reported on the top-left corner of each error map. Red pixels on the error map represent areas where the error is higher than the threshold, i.e., 0.07. It is clear that our method has less error as compared to previous methods, especially in the occlusion and edge regions. }
  \label{fig: comp}
\end{figure*}

\begin{figure}[ht]
  \centering
  \includegraphics[width= \linewidth]{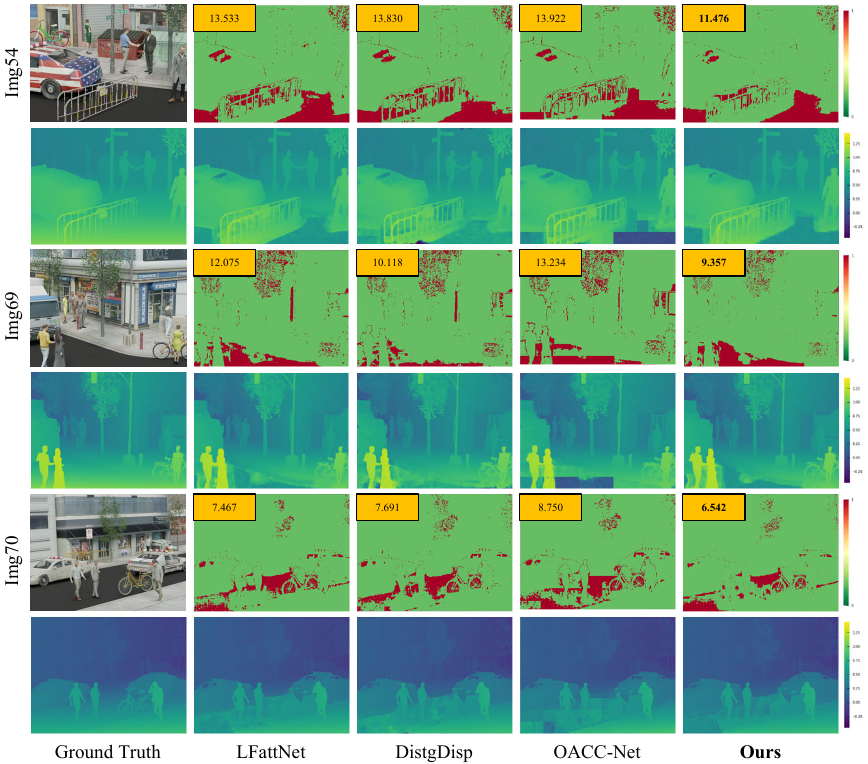}

  \caption{Visual comparisons between our method and state-of-the-art methods on the scenes of UrbanLF-Syn dataset\cite{sheng2022urbanlf}, i.e., \textit{Img27}, \textit{Img54}, \textit{Img69} and \textit{Img70}, including LFattNet \cite{tsai2020attention}, DistgDisp \cite{wang2022disentangling}, OACC-Net \cite{wang2022occlusion}, with the corresponding BadPix 0.07 error maps. Lower is better. The best results are shown in boldface. Please zoom in for a better comparison.
  }
  \label{fig: compare_urban}
\end{figure}

\begin{figure}[ht]
  \centering
  \includegraphics[width= \linewidth]{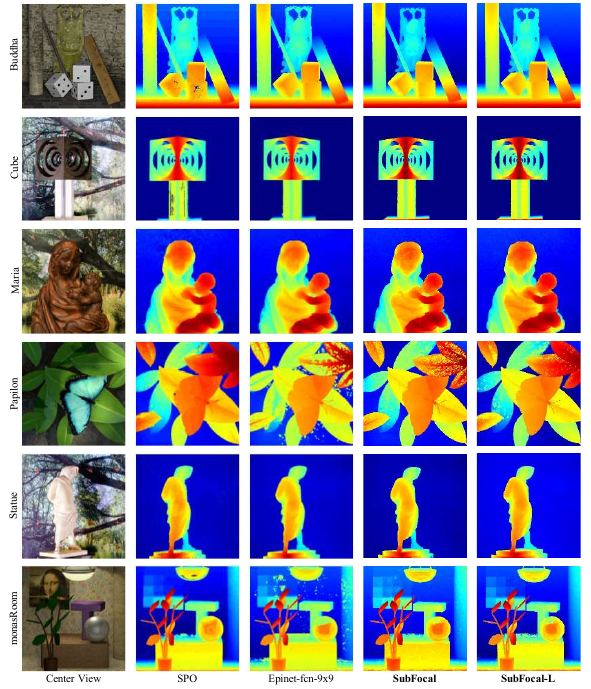}
  \caption{Visual results achieved by SPO \cite{zhang2016robust}, EPINET \cite{shin2018epinet}, 
  and our method on the old HCI LF dataset \cite{vaish2008new}.}
  \label{fig: real_3}
\end{figure}

\begin{figure}[!ht]
  \centering
  \includegraphics[width=\linewidth]{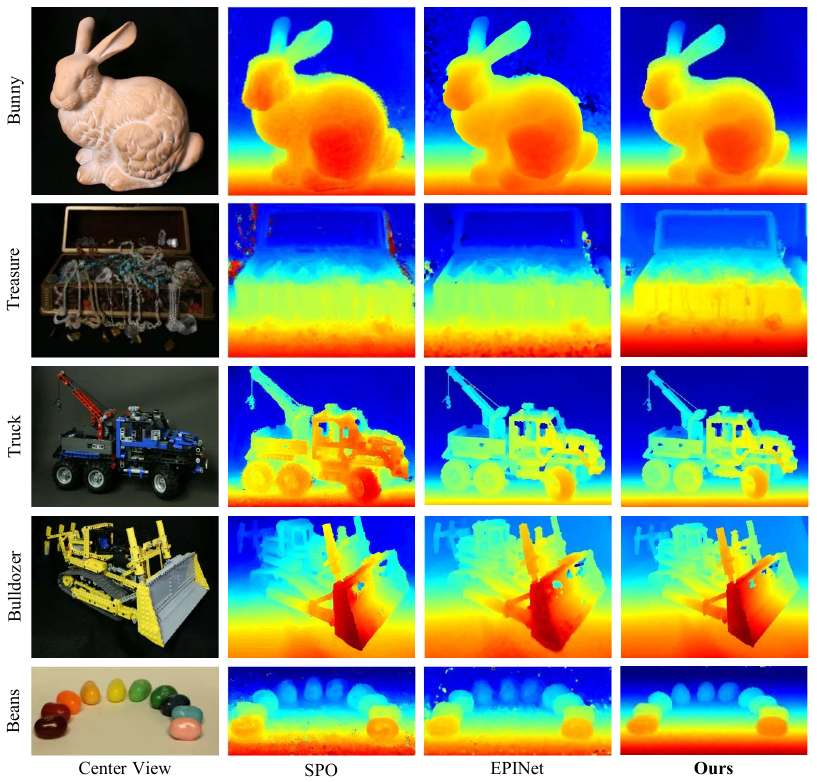}

    \caption{Evaluation on real-world scenes. Compared to SPO \cite{zhang2016robust}, EPINET \cite{shin2018epinet}, our disparity maps show better performance, even in complex scenes, e.g., \textit{Truck}, \textit{Bulldozer}. }
  \label{fig: real}
\end{figure}

\subsection{Comparison with the state-of-the-art methods}

\subsubsection{Quantitative Comparison}
We performed quantitative comparison with the state-of-the-art methods, among which PS\_RF \cite{jeon2018depth} is the MVS-based method, SPO \cite{zhang2016robust} and EPI\_ORM \cite{li2020epi} are the EPI-based methods,  CAE \cite{park2017robust}, OBER-cross-ANP \cite{schilling2018trust} and OAVC \cite{han2021novel} are the Defocus-based methods, Epinet-fcn-m \cite{shin2018epinet}, LFattNet \cite{tsai2020attention}, AttMLFNet \cite{chen2021attention}, FastLFNet \cite{huang2021fast}, DistgDisp \cite{wang2022disentangling} and OACC-Net \cite{wang2022occlusion} are the DL-based methods. As demonstrated in Fig. \ref{fig: screenshot} our method ranks first place on all the four commonly used metrics (BadPix 0.01, BadPix 0.03, BadPix 0.07, and MSE $\times$100) on the HCI 4D LF benchmark. Table \ref{table: quantitative} shows the quantitative comparison results on the validation and test sets. Our method achieves the best or second-best performance in most scenes, especially on the unlabeled test sets, and also achieves an overall best performance, as shown in Table \ref{table: avg}. SubFocal surpasses the state-of-the-art method (i.e., AttMLFNet \cite{chen2021attention}) by \textbf{8.3\%} (from 16.42 to 15.06) on BadPix 0.01, \textbf{14.7\%} (from 6.568 to 5.602) on BadPix 0.03, \textbf{17.7\%} (from 3.596 to 2.956)on BadPix 0.07, and \textbf{5.9\%} (from 1.720 to 1.618) on MSE $\times$100, respectively. Furthermore, SubFocal-L (the enhanced version) achieves a \textbf{21.7\%} (from 16.42 to 12.85) gain on BadPix 0.01, \textbf{28.5\%} (from 6.568 to 4.697) gain on BadPix 0.03, \textbf{23.9\%} (from 3.596 to 2.735) gain on BadPix 0.07, and \textbf{8.1\%} (from 1.720 to 1.581) gain on MSE $\times$100, respectively. We also have compared our method with three state-of-the-art methods on the UrbanLF-Syn dataset\cite{sheng2022urbanlf}, including LFattNet \cite{tsai2020attention}, DistgDisp \cite{wang2022disentangling}, OACC-Net \cite{wang2022occlusion}. For fair comparisons, we retrained these methods based on released codes and adjusted disparity ranges, respectively. The ranges of LFattNet and DistgDisp were set to [-1, 2], and OACC-Net was [-2, 2]. The disparity interval was set to 1. Table \ref{table: quantitative_urban} shows the comparison results on the UrbanLF-Syn dataset. Our method ranks first in most scenes and achieves the top average result in terms of BadPix 0.07, outperforming current state-of-the-art methods.

\subsubsection{Qualitative Comparison} 
Figure \ref{fig: comp} shows the qualitative results achieved by different methods on four scenes of the HCI 4D benchmark: \textit{Backgammon}, \textit{Dots}, \textit{Stripes}, and \textit{Boxes}. Scenes \textit{Backgammon} and \textit{Stripes} are stratified and contain multiple lines. Scene \textit{Dots} is composed of regular dots with clear edges. Scene \textit{Boxes} consists of multiple objects and contains severe occlusions. The BadPix 0.07 error maps show that our approach has substantially less error as compared with previous methods, especially in the occlusion and edge regions. Figure \ref{fig: compare_urban} shows qualitative comparison results on \textit{Img27}, \textit{Img54}, \textit{Img69}, and \textit{Img70} scenes of  Urban-Syn dataset. From this example, our method can handle the edge regions better than other methods.
We attribute our improved performance to two factors. Firstly, our sub-pixel cost volume produces a more refined disparity distribution compared to previous methods. Secondly, our uncertainty-aware focal loss assigns higher weights to difficult points in the edge regions by computing the disparity distribution difference, leading to even better results. Therefore, our approach is able to produce a more reasonable disparity distribution and accurate disparity map.

\subsubsection{Generalization Comparison}
In order to validate the generalization ability of our method, we compare it with the mainstream methods in various scenes. We employ the model trained on the HCI dataset directly for inference. First, we compare the visual results achieved by SPO \cite{zhang2016robust}, EPINET \cite{shin2018epinet}, 
  and our method on the old HCI LF dataset \cite{vaish2008new}. As shown in Fig.~\ref{fig: real_3}, our results exhibit sharp edges and less noise in comparison with other methods, such as the scene \textit{Papilon} and \textit{monasRoom}. We also evaluate our method on the real-world LF images taken with a Lytro Illum camera \cite{bok2016geometric} and a moving camera mounted on a gantry \cite{vaish2008new}. Fig. \ref{fig: real} illustrates that our method can produce more accurate disparity maps than SPO \cite{zhang2016robust} and EPINet \cite{shin2018epinet}, such as the boom of the scene \textit{Truck} and the background of the scene \textit{Dinosaur}. It clearly demonstrates the superior generalization capability of our approach. 

\begin{table}[tb]
\centering
\caption{Comparative results achieved by our method with different interpolation methods and disparity sampling density. The inference with an input LF image of 9$\times$9$\times$512$\times$512 is performed on an NVIDIA RTX 3090 GPU.  $\dagger $ means to perform patch cropping for inference.}
\renewcommand\arraystretch{1.0}
\resizebox{\columnwidth}{!}{

\begin{tabular}{c|ccccc}
\toprule
Interv. & Interp. & BP 0.07 & MSE & Training Time & Test Time \\ 
\midrule
0.5 & phase & 2.44 & 1.008 & 100h & 24.0s$^\dagger $      \\ 
0.5 & nearest  & 2.20 & 0.925 & \textbf{50h} & \textbf{3.0s}     \\
0.5 & bilinear & \textbf{2.04} & \textbf{0.845} & 54h & 5.7s  \\

\midrule
1 & bilinear & 2.79 & 1.245 & \textbf{39h} & \textbf{3.5s}   \\ 
0.5 & bilinear & 2.04 & 0.845 & 54h & 5.7s  \\
0.25 & bilinear & 1.92 & 0.839 & 110h & 18.7s      \\
0.1 & bilinear & \textbf{1.73} & \textbf{0.748} & 330h & 60.0s$^\dagger $      \\
\bottomrule

\end{tabular}
}

\label{table: sub}
\end{table}

\begin{figure}[tb]
  \centering
  \includegraphics[width=\linewidth]{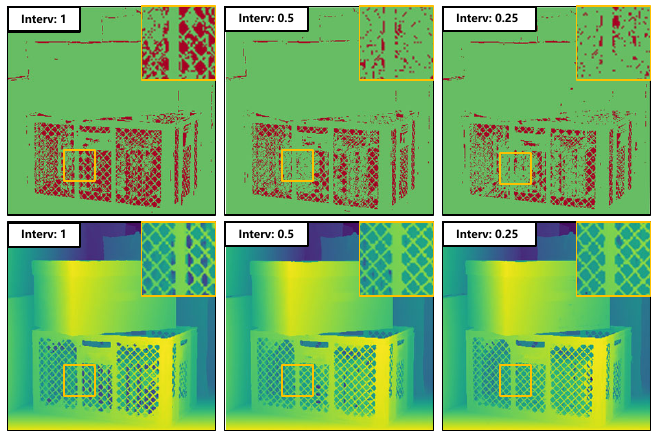}

  \caption{Visual comparison of the results and BadPix 0.07 error maps of different disparity intervals on \textit{Boxes} scene.}
  \label{fig: sub}
\end{figure}

\subsection{Ablation Study}
\label{subsec:ablation}
We conduct extensive ablation experiments to analyze the effectiveness of our method. Our ablation studies include the interpolation approaches, disparity intervals, combinations of loss functions, hyper-parameters of UAFL, and visualization of uncertainty maps.

\subsubsection{Interpolation Methods }
\label{ablation: inter}

We compare three representative interpolation methods: phase\cite{jeon2015accurate}, nearest neighbor, and bilinear. As shown in the upper part of \ref{table: sub}, the method of phase shift requires more training and testing time, while the result is inferior to the nearest neighbor and bilinear approaches. 
The bilinear approach improves the results over the nearest approach (0.02 in terms of MSE) while taking a little more time in training and testing. Therefore, we finally choose the bilinear approach to construct a sub-pixel cost volume.

\subsubsection{Disparity Interval}
\label{ablation: interval}

The selection of the disparity interval determines the sampling density and further affects the inference speed of the model. We need to choose the proper disparity interval to achieve a good trade-off in accuracy and speed. We set the disparity intervals to 1, 0.5, 0.25, and 0.1, respectively, and then train our model for 50 epochs on the HCI 4D LF benchmark and then evaluate their performance on the validation set. 
The lower part of Table \ref{table: sub} shows the results of the model with different disparity intervals. As the disparity interval decreases and the number of samples increases, the accuracy of the model is improved, while the training time increases as well. 
Specifically, reducing the disparity interval from 1 to 0.5 improves the BadPix 0.07 and MSE $\times$100 by 0.75 and 0.4, respectively. 
Fig. \ref{fig: sub} illustrates that the disparity map at the interval of 0.5 is more accurate than the interval of 1 on the heavily occluded region. In addition, we find no significant visual difference between the results of disparity intervals 0.25 and 0.5. Therefore, based on the consideration of accuracy and speed, we select a disparity interval of 0.5 in our experiments.

\begin{table}[tb]
\centering
\caption{Quantitative comparison of the results at different combinations of loss functions.}
\label{table: loss}
\renewcommand\arraystretch{1.0}
\resizebox{\columnwidth}{!}{

\begin{tabular}{cccccc|c}
\toprule
baseline & L1 & MSE & JS & KL & UAFL & BadPix 0.07 \\ 
\hline
$\checkmark$ &  &  &  &  &  & 2.04 \\ \hline
&  & $\checkmark$ &  &  &  & 2.64\\ \hline
&  &  & $\checkmark$ &  &  & 2.08\\ \hline
& $\checkmark$ &  & $\checkmark$ &  &  & 1.94 \\ \hline
& $\checkmark$ &  &  & $\checkmark$ &  & 2.36\\ \hline
&  &  &  &  & $\checkmark$& \textbf{1.93}\\ 
\bottomrule
\end{tabular}

}

\end{table}

\begin{figure}[tb]
  \centering
  \includegraphics[width=\linewidth]{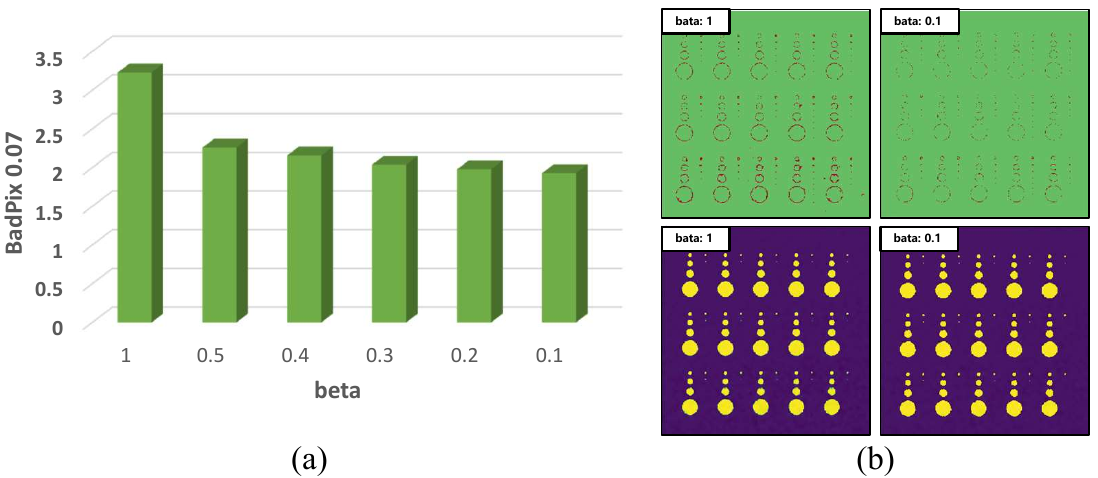}

  \caption{Results achieved by our method with different $\beta$. (a) BadPix 0.07 metric for $\beta$ decreased from 1 to 0.1. (b) Disparity maps and BadPix 0.07 error maps with $\beta=1$ and $\beta=0.1$ on \textit{Dots} scene.}
  \label{fig: beta}
\end{figure}

\subsubsection{Combinations of Loss Functions}

Selecting appropriate loss functions is a crucial aspect of our method. We test some combinations of common losses, including L1 loss, MSE loss, KL divergence loss, and JS divergence loss. 
We use the model with a disparity interval of 0.5 as the pre-trained baseline, and its network structure adopts LFattNet \cite{tsai2020attention}. Then the model is finetuned for 10 epochs with the learning rate being set to 1$\times$10$^{-4}$. Table \ref{table: loss} shows the results of different loss functions. Our UAFL achieves the lowest BadPix 0.07 metrics.

\begin{figure}[tb]
  \centering
  \includegraphics[width=\linewidth]{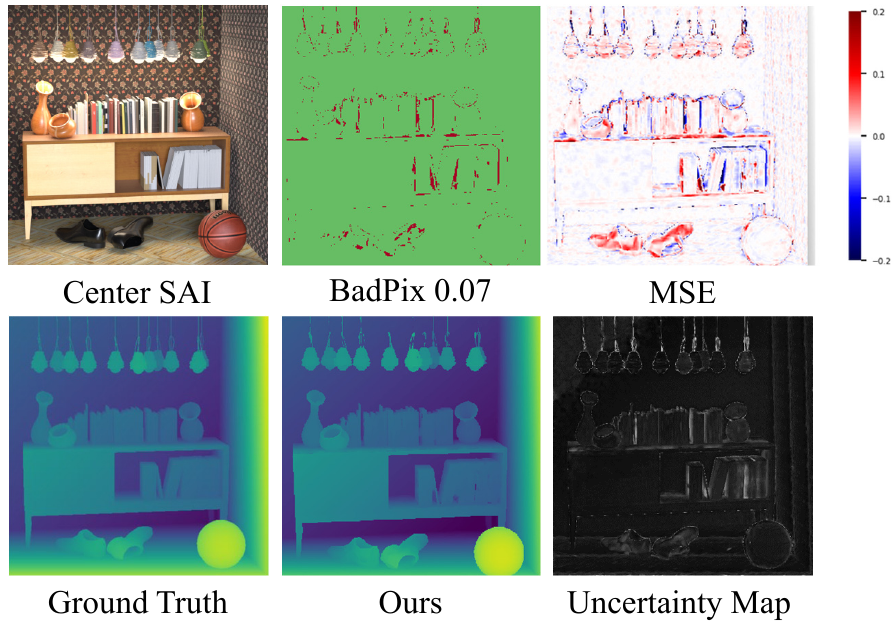}
  \caption{Qualitative result on the $Dino$ scene. In an uncertainty map, black colors mean small uncertainty while white colors denote high uncertainty.}
  \label{fig: focal}
\end{figure}

\begin{figure}[tb]
  \centering
  \includegraphics[width=\linewidth]{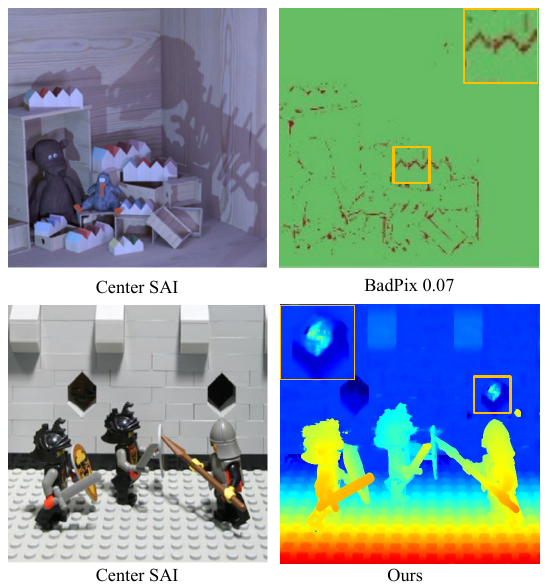}

  \caption{Error analysis of the \textit{Dino} scene and visualization of the result of the \textit{Knight} scene.}
  \label{fig: limit}
\end{figure}

\subsubsection{Hyper-parameter of UAFL}
\label{ablation: beta}

The function of UAFL is to dynamically assign weights based on an uncertainty map to supervise the training of disparity distribution and disparity simultaneously. We introduce the hyper-parameter $\beta$ to adjust the ratio of the assigned weights. Based on the disparity interval of 0.5, we choose different $\beta$ to investigate its influence, and the results are shown in Fig. \ref{fig: beta}(a). We observe that as $\beta$ decreases from 1 to 0.1, the BadPix 0.07 metric is improved. Fig. \ref{fig: beta}(b) also shows that the BadPix 0.07 error map with $\beta=0.1$ is significantly better than that with $\beta=1$. Finally, we set $\beta = 0.1$ in our model.

\subsubsection{Visualization of Uncertainty Map}

Fig. \ref{fig: focal} shows the per-pixel qualitative result on the \textit{Dino} scene,
which contains several hard regions, e.g., heavy occlusions and sharp edges. 
We can observe from the uncertainty map that our method assigns higher weights to the hard regions and lower weights to easy regions. Therefore, our approach can improve the overall accuracy of disparity estimation.

\section{Limitations and Discussion}
\label{sec:limit} 
Although our method performs well in most scenes, we do not explicitly consider occlusions and texture-less regions. As shown in Fig. \ref{fig: limit}, relatively large errors occur in occlusion regions (see \textit{Dino} scene error map), and artifacts may appear in texture-less regions (see the black hole in \textit{Knight} scene). We also verified the effect of the sub-pixel cost volume. As shown in Table \ref{table: sub} and Table \ref{table: quantitative_urban}, a small sampling interval results in increased computational cost and longer inference time, making it unfeasible for practical applications. Therefore, it is wise to make a trade-off between accuracy and speed based on specific requirements. In future work, we plan to enhance the robustness of our method to handle challenging scenarios such as non-Lambertian surfaces, weak textures, and highlight regions. Additionally, inspired by OACC-Net \cite{wang2022occlusion}, we plan to explore the construction of cost volume based on dilated convolution to reduce inference time.

\section{Conclusion} 
\label{sec:conclusion} 
In this paper, we have presented a straightforward and efficient approach to learning the disparity distribution for LF depth estimation. On one hand, we have constructed an interpolation-based cost volume at the sub-pixel level for a finer disparity distribution. On the other hand, we have designed an uncertainty-aware focal loss based on JS divergence to supervise the disparity distribution. Extensive experiments validate the effectiveness of our method. Our method outperforms state-of-the-art methods, as demonstrated by our experimental results. Our method holds great potential for advancing the field of LF-based depth estimation research.


 

\bibliographystyle{IEEEtran}
\bibliography{reference}


 



\begin{IEEEbiography}[{\includegraphics[width=1in,height=1.25in,clip,keepaspectratio]{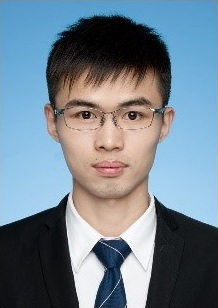}}]{Wentao Chao}
received his B.E. degree in the College of Automation Engineering from Shanghai University of Electric Power, Shanghai, China, in 2017, and his M.S. degree in the Department of Automation from North China Electric Power University, Baoding, China, in 2020. He is now a Ph.D. candidate at the School of Artificial Intelligence, Beijing Normal University, Beijing, China. His research interests include generative models, image processing, light field imaging, and 3D reconstruction.
\end{IEEEbiography}

\begin{IEEEbiography}[{\includegraphics[width=1in,height=1.25in,clip,keepaspectratio]{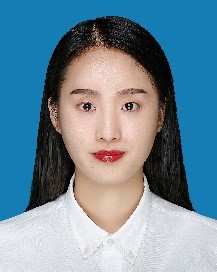}}]{Xuechun Wang}
received a B.S. degree in information and computing science from Yunnan University, Kunming, China, in 2017, and an M.S. degree in computing mathematics from Yunnan University, Kunming, China, in 2020. Currently, she is pursuing a Ph.D. degree in the School of Artificial Intelligence at Beijing Normal University, Beijing, China. Her research interests include camera calibration, light field imaging, image processing, and 3D reconstruction.
\end{IEEEbiography}

\begin{IEEEbiography}[{\includegraphics[width=1in,height=1.25in,clip,keepaspectratio]{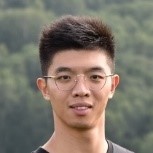}}]{Yingqian Wang}
received his B.E. degree in electrical engineering from Shandong University, Jinan, China, in 2016, the Master and the Ph.D. degrees in information and communication engineering from National University of Defense Technology (NUDT), Changsha, China, in 2018 and 2023, respectively. He is currently an assistant professor with the College of Electronic Science and Technology, NUDT. His research interests focus on optical imaging and detection, particularly on light field imaging, image super-resolution and infrared small target detection.
\end{IEEEbiography}

\begin{IEEEbiography}[{\includegraphics[width=1in,height=1.25in,clip,keepaspectratio]{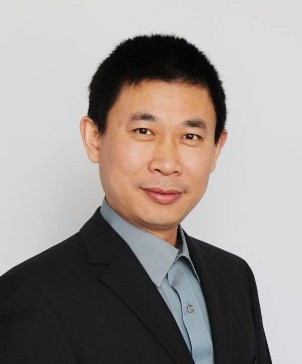}}]{Guanghui Wang}
received the Ph.D. degree in engineering from University of Waterloo, Waterloo, ON, Canada, in 2014. He is currently an Associate Professor with the Department of Computer Science, Toronto Metropolitan University, Toronto, ON, Canada. He was with the University of Kansas, Lawrence, KS, USA, as an Assistant Professor and then Associate Professor, from 2014 to 2020. He has authored one book "Guide to Three Dimensional Structure and Motion Factorization" (Springer-Verlag). He has authored or coauthored over 180 papers in peer reviewed journals and conferences. His research interests include computer vision, image analysis, machine learning, and intelligent systems.
\end{IEEEbiography}

\begin{IEEEbiography}[{\includegraphics[width=1in,height=1.25in,clip,keepaspectratio]{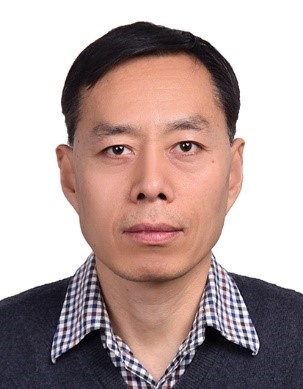}}]{Fuqing Duan}
received the B.S. and M.S. degrees in mathematics from Northwest University, Xi’an, China, in 1995 and 1998, respectively, and the Ph.D. degree in pattern recognition from the National Laboratory of Pattern Recognition, Beijing, China, in 2006. He is currently a Professor with the School of Artificial Intelligence, Beijing Normal University, Beijing. His current research interests include 3D reconstruction, skull identification, and machine learning and applications. He has authored more than 80 conference and journal articles on related topics.
\end{IEEEbiography}




\end{document}


\title{Learning Sub-Pixel Disparity Distribution for Light Field Depth Estimation\\(Supplementary Material)}

\author{Wentao Chao, Xuechun Wang, Yingqian Wang, Guanghui Wang,~\IEEEmembership{Senior Member,~IEEE}, Fuqing Duan}



\maketitle

\section{Appendix}

Our SubFocal is described in detail in Sec. \ref{sup:network}. The 4D light field (LF) benchmark is further compared in Sec. \ref{sub:results}. Section \ref{sub:visual} displays additional visual results obtained using various techniques on other LF datasets \cite{le2018light,rerabek2016new,vaish2008new,wanner2013datasets}. Section \ref{sub:future} discusses the future direction of our approach.

\subsection{Details of our SubFocal}
\label{sup:network}

\begin{figure}[htb]
  \centering
  \includegraphics[width=\linewidth]{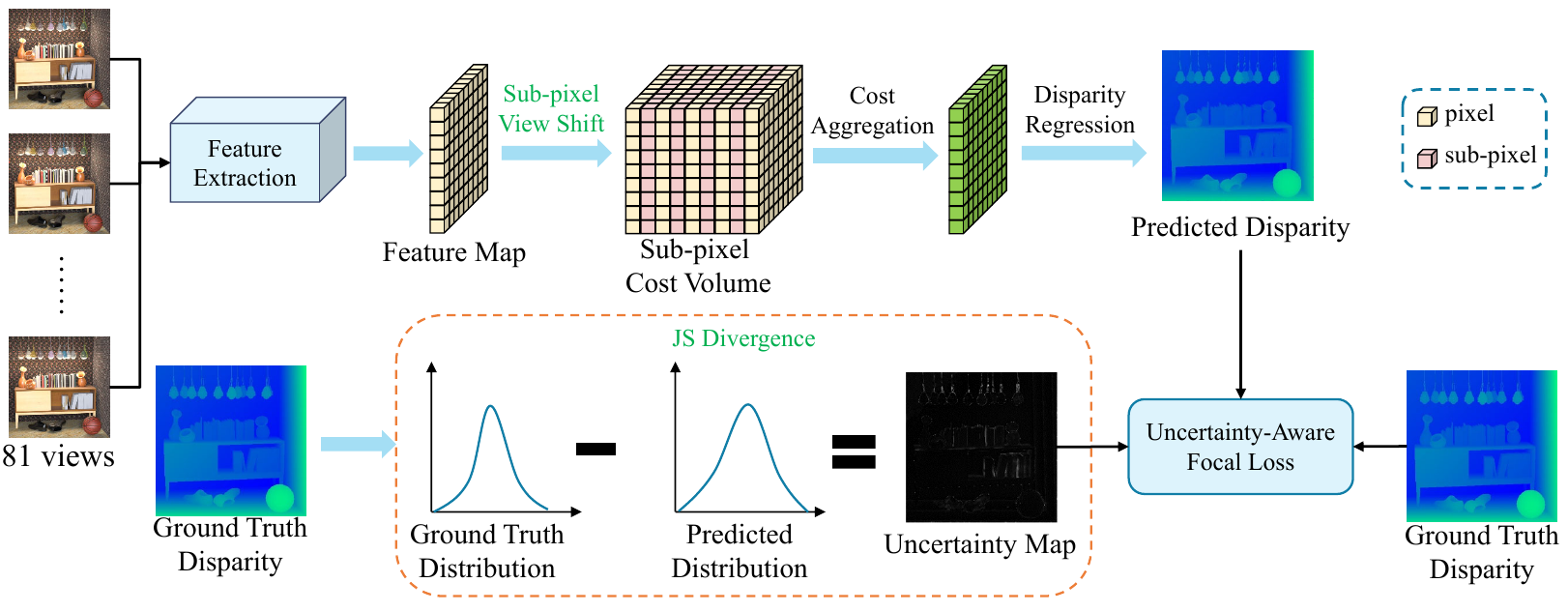}

  \caption{The architecture of our SubFocal network, including feature extraction, sub-pixel cost volume construction, cost aggregation, and disparity regression.}
  \label{fig: network2}
\end{figure}

\begin{table}[tb]
\centering
\caption{The detailed architecture of our SubFocal. 
$Conv2D$, $Conv3D$, $ResBlock3D$ represents 2D convolution, 3D convolution and 3D residual block, respectively. $H$ and $W$ are the height and width. $M$ denotes the number of views (e.g.,$M=9\times9=81$), and $D$ denotes the number of disparity candidates (e.g.,$D=17$).}
\label{table: network}
\renewcommand\arraystretch{1.5}
\resizebox{\columnwidth}{!}{
\begin{tabular}{|cccc|}
\hline
\multicolumn{1}{|c|}{Layers} & \multicolumn{1}{c|}{Kernel Size} & \multicolumn{1}{c|}{Input Size} & Output Size \\ \hline
\multicolumn{4}{|c|}{Feature Extraction} \\ \hline
\multicolumn{1}{|c|}{Conv2D\_1} & \multicolumn{1}{c|}{3$\times$3} & \multicolumn{1}{c|}{M$\times$(H$\times$W$\times$1)} & M$\times$(H$\times$W$\times$4) \\ \hline
\multicolumn{1}{|c|}{Conv2D\_2} & \multicolumn{1}{c|}{3$\times$3} & \multicolumn{1}{c|}{M$\times$(H$\times$W$\times$4)} & M$\times$(H$\times$W$\times$4) \\ \hline
\multicolumn{1}{|c|}{SPP Module} & \multicolumn{1}{c|}{\begin{tabular}[c]{@{}c@{}}2$\times$2\\ 4$\times$4\\ 8$\times$8\\ 16$\times$16\end{tabular}} & \multicolumn{1}{c|}{M$\times$(H$\times$W$\times$4)} & M$\times$(H$\times$W$\times$4) \\ \hline
\multicolumn{4}{|c|}{Sub-pixel Cost Volume Construction} \\ \hline
\multicolumn{1}{|c|}{Shift\&Concat} & \multicolumn{1}{c|}{-} & \multicolumn{1}{c|}{M$\times$(H$\times$W$\times$4)} & D$\times$H$\times$W$\times$(4$\times$M) \\ \hline
\multicolumn{1}{|c|}{Channel Attention} & \multicolumn{1}{c|}{1x1x1} & \multicolumn{1}{c|}{D$\times$H$\times$W$\times$(4$\times$M)} & D$\times$H$\times$W$\times$(4$\times$M) \\ \hline
\multicolumn{4}{|c|}{Cost Aggregation} \\ \hline
\multicolumn{1}{|c|}{Conv3D\_1} & \multicolumn{1}{c|}{3$\times$3$\times$3} & \multicolumn{1}{c|}{D$\times$H$\times$W$\times$(4$\times$M)} & DxH$\times$W$\times$150 \\ \hline
\multicolumn{1}{|c|}{Conv3D\_2} & \multicolumn{1}{c|}{3$\times$3$\times$3} & \multicolumn{1}{c|}{D$\times$H$\times$W$\times$150} & D$\times$H$\times$W$\times$150 \\ \hline
\multicolumn{1}{|c|}{ResBlock3D $\times$2} & \multicolumn{1}{c|}{\begin{tabular}[c]{@{}c@{}}3$\times$3$\times$3\\ 3$\times$3x3\end{tabular}} & \multicolumn{1}{c|}{D$\times$H$\times$W$\times$150} & D$\times$H$\times$W$\times$150 \\ \hline
\multicolumn{1}{|c|}{Conv3D\_3} & \multicolumn{1}{c|}{3$\times$3$\times$3} & \multicolumn{1}{c|}{D$\times$H$\times$W$\times$150} & D$\times$H$\times$W$\times$150 \\ \hline
\multicolumn{1}{|c|}{Cost} & \multicolumn{1}{c|}{3$\times$3$\times$3} & \multicolumn{1}{c|}{D$\times$H$\times$W$\times$150} & D$\times$H$\times$W$\times$1 \\ \hline
\multicolumn{1}{|c|}{Squeeze\&Transpose} & \multicolumn{1}{c|}{-} & \multicolumn{1}{c|}{D$\times$H$\times$W$\times$1} & H$\times$W$\times$D \\ \hline
\multicolumn{4}{|c|}{Disparity Regression} \\ \hline
\multicolumn{1}{|c|}{Softmax} & \multicolumn{1}{c|}{-} & \multicolumn{1}{c|}{H$\times$W$\times$D} & H$\times$W$\times$D \\ \hline
\multicolumn{1}{|c|}{Regress} & \multicolumn{1}{c|}{-} & \multicolumn{1}{c|}{H$\times$W$\times$D} & H$\times$W$\times$1 \\ \hline
\end{tabular}
}

\end{table}

Fig. \ref{fig: network2} shows the pipeline of our SubFocal, including feature extraction, sub-pixel cost volume construction, cost aggregation, and disparity regression. The detailed structure of our SubFocal is shown in Table \ref{table: network}. We describe the details of each module in detail below.

\subsubsection{Feature Extraction} 
First, two 3 $\times$ 3 convolutions (i.e., \textit{Conv2D\_1} and \textit{Conv2D\_2)} are used to extract the initial feature with a channel of 4. Then, we use the SPP module to extract multi-scale features. SPP module is set as follows:
\begin{enumerate}

\item  Four average pooling operations at different scales are used to compress the features. The sizes of the average pooling blocks are 2 $\times$ 2, 4 $\times$ 4, 8 $\times$ 8, and 16 $\times$ 16. 
\item  A 1 $\times$ 1 convolution layer is used for reducing the feature dimension for each scale. 
\item  Bilinear interpolation is adopted to upsample these low-dimensional feature maps to the same size.
\item  Concatenating the feature maps of all levels as the output feature map of the SPP module.

\end{enumerate}

\subsubsection{Sub-pixel Cost Volume Construction}

\textit{Shift-and-Concat} operation\cite{tsai2020attention} is used to construct sub-pixel cost Volume. We manually shift the input images along the $u$ or  $v$ direction with different disparity levels. In our setting, we have 17 disparity levels ranging from -4 to 4, where the sub-pixel interval is 0.5. After shifting the feature maps, we concatenate these feature maps into a 4D cost volume $D\times H\times W \times (4\times M)$. 

Additionally, we dynamically weigh the cost volume using channel attention. The channel module produces the attention map by a global pooling layer, then two fully connected layers, and finally a sigmoid layer. To make the attended features, we then use the element-wise product to multiply the features from the cost volume by the corresponding attention scores.

\subsubsection{Cost Aggregation}

Our architecture consists of eight $3\times3\times3$ convolutional layers, with two residual blocks from the third to the sixth 3D convolutional layers. Then We use the squeeze and transpose operation to adjust dimensions. Finally,  the output cost volume of cost aggregation is 3D tensor $H\times W\times D$. 

\subsubsection{Disparity Regression}
We use the softmax operation to calculate the disparity distribution $H\times W\times D$. Then, the final disparity map $H\times W\times 1$ is calculated by the weighted sum of each disparity distribution with its normalized probability as the weight.

\subsection{Results on the 4D LF Benchmark}
\label{sub:results}
\begin{figure*}[htb]
  \centering
  \includegraphics[width=0.95 \linewidth]{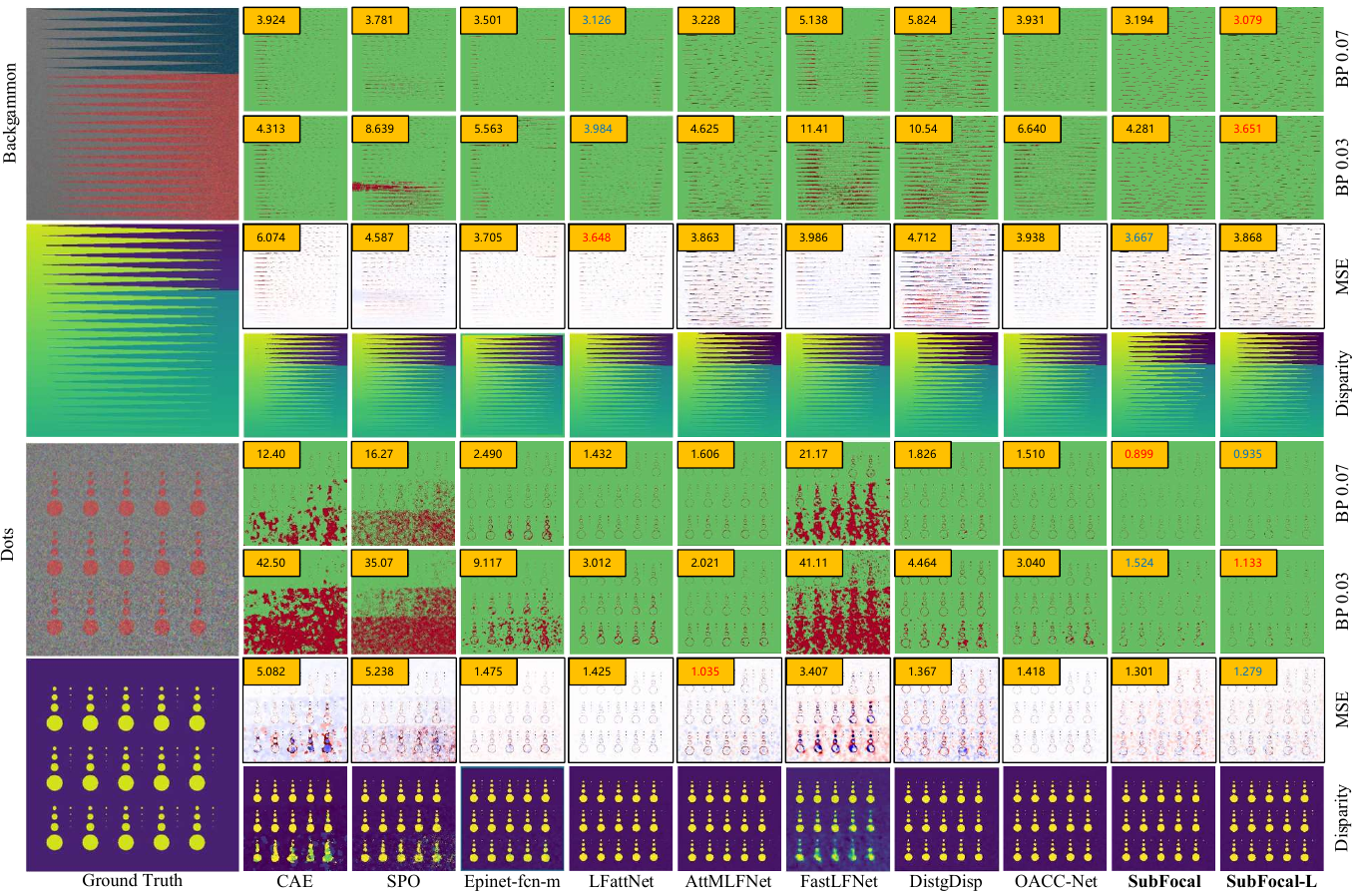}
  \includegraphics[width=0.95 \linewidth]{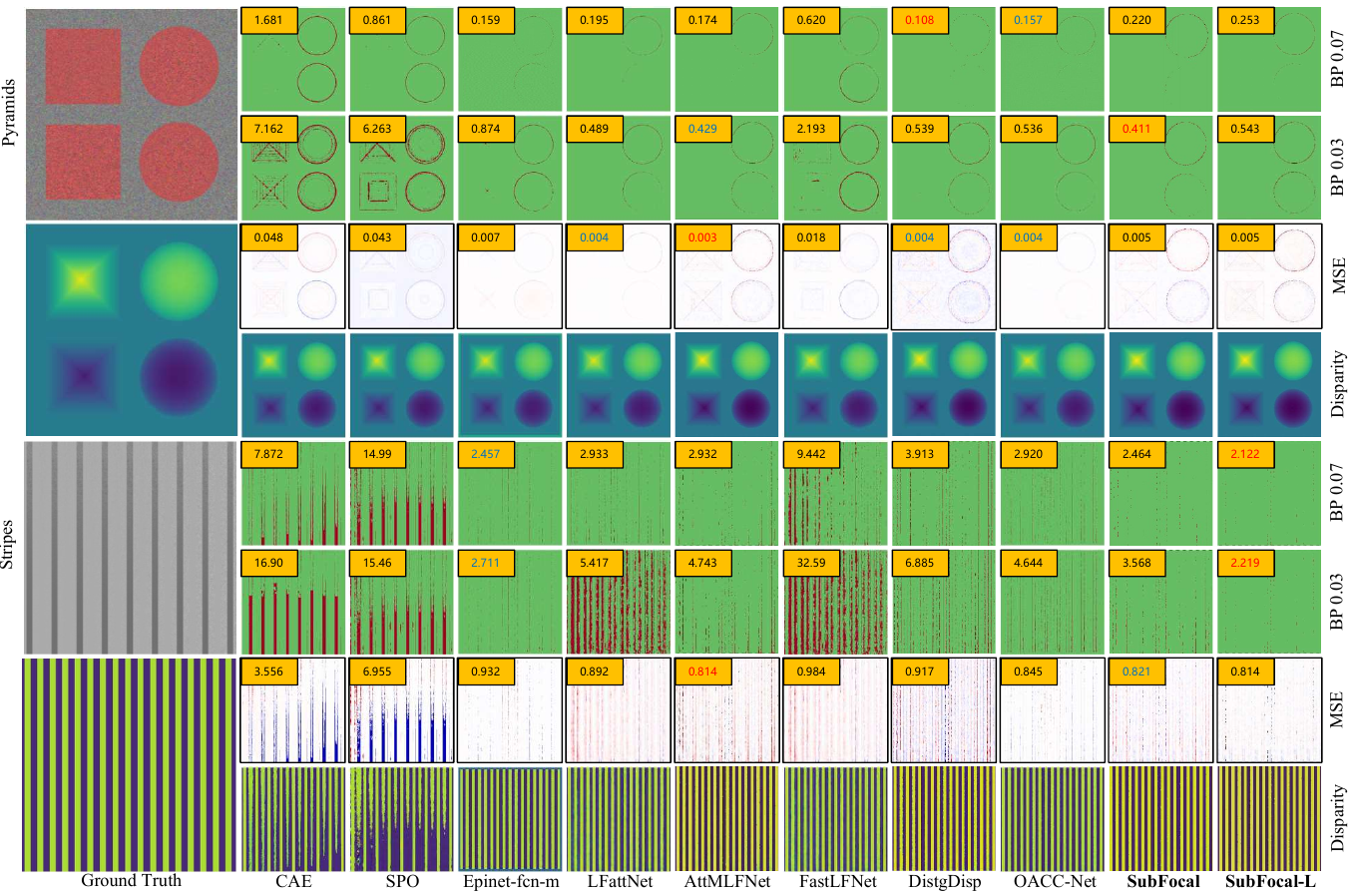}
  \caption{Visual comparisons of disparity and error maps on validation scenes \textit{backgammon}, \textit{dots}, \textit{pyramids}, and \textit{stripes} \cite{rerabek2016new}. Corresponding quantitative scores (BadPix0.07, BadPix0.03, and MSE) are reported on the top-left corner of each error map.
  }

  \label{fig: hci_1}
\end{figure*}

\begin{figure*}[htb]
  \centering
  \includegraphics[width=0.95\linewidth]{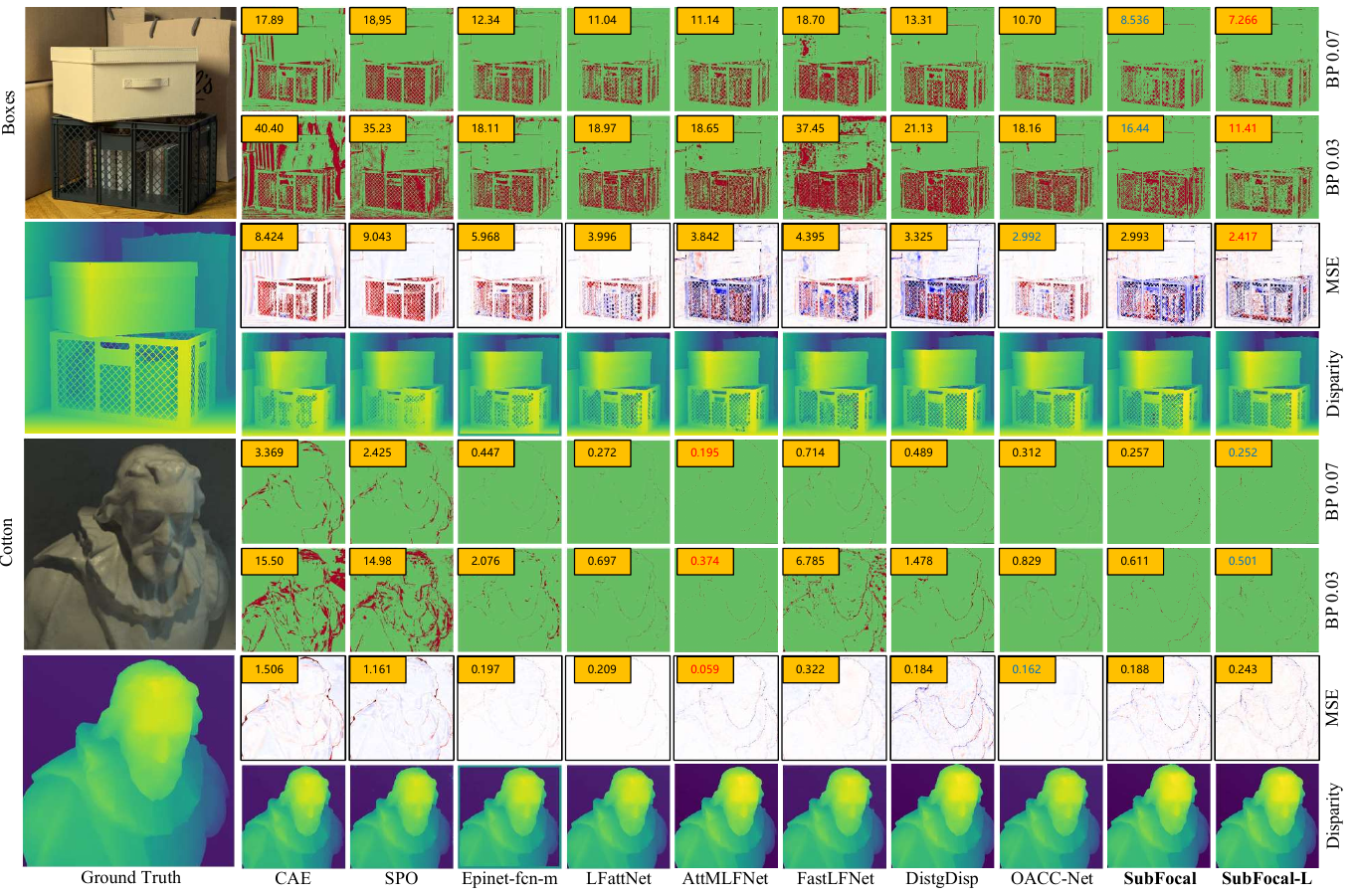}

  \includegraphics[width=0.95\linewidth]{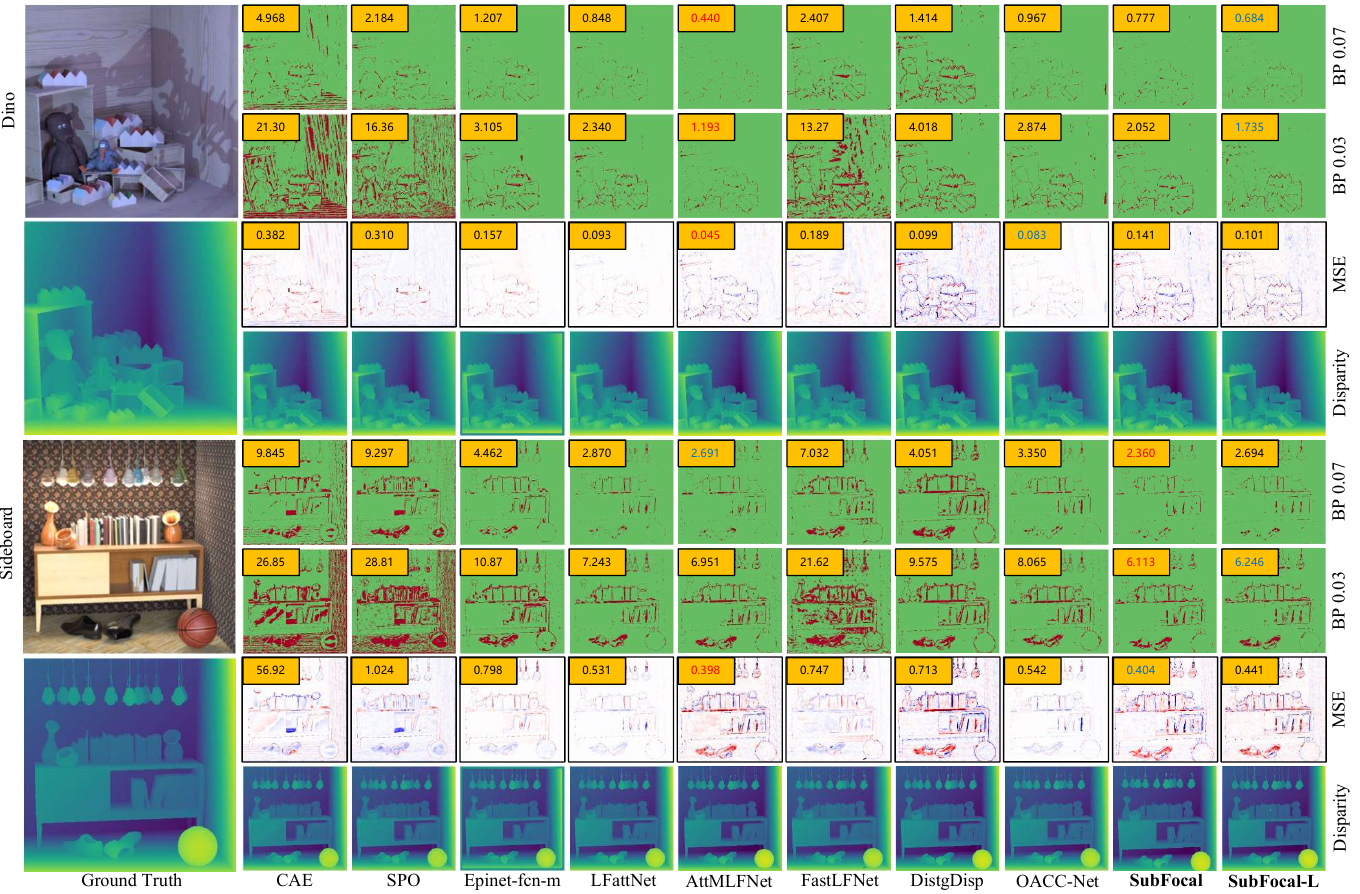}

  \caption{Visual comparisons of disparity and error maps on validation scenes \textit{boxes}, \textit{cotton}, \textit{dino}, and \textit{sideboard} \cite{rerabek2016new}. Corresponding quantitative scores (BadPix0.07, BadPix0.03, and MSE) are reported on the top-left corner of each error map.
  }

  \label{fig: hci_2}
\end{figure*}

\begin{figure*}[htb]
  \centering

  \includegraphics[width=\linewidth]{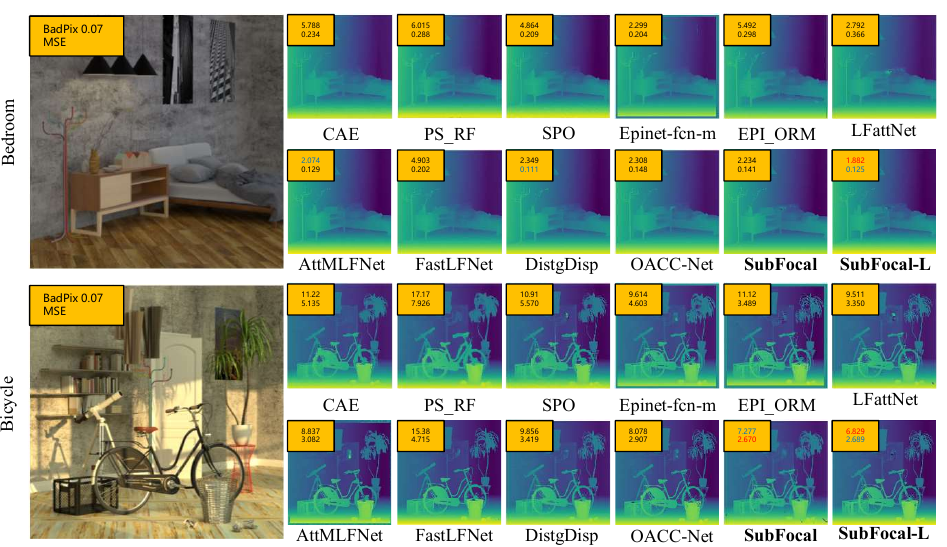}

  \includegraphics[width= \linewidth]{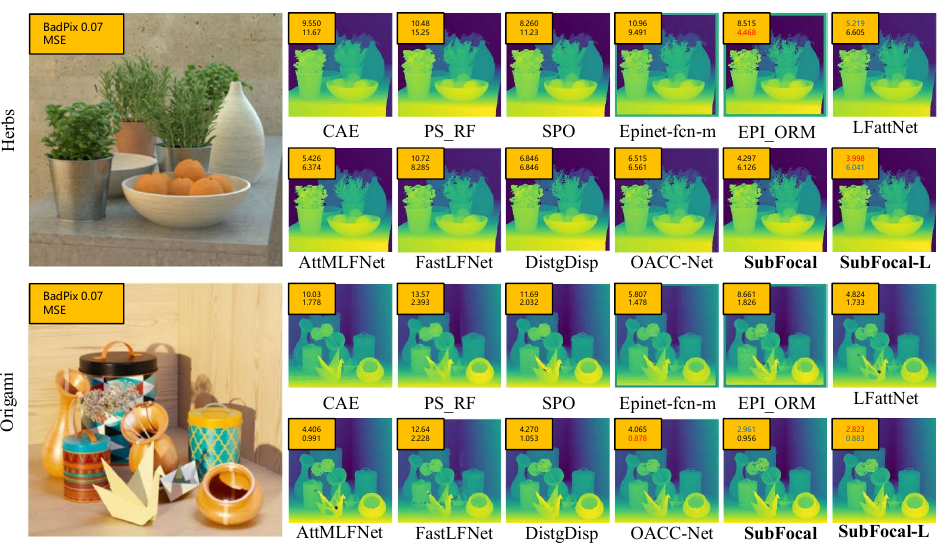}

  \caption{Visual comparisons of disparity maps on test scenes \textit{bedroom}, \textit{bicycle}, \textit{herbs}, and \textit{origami} \cite{rerabek2016new}. The ground-truth disparity maps of these scenes are not released. The BadPix 0.07 and MSE of each method (copied from the benchmark site) are reported on the left-top corner.
  }

  \label{fig: hci_3}
\end{figure*}

Fig. \ref{fig: hci_1} and Fig. \ref{fig: hci_2} depict the estimated disparity maps and error maps for the eight validation scenes. The estimated disparity maps for the four test scenes are shown in Fig. \ref{fig: hci_3}.

\begin{figure*}[htb]
  \centering
  \includegraphics[width=\linewidth]{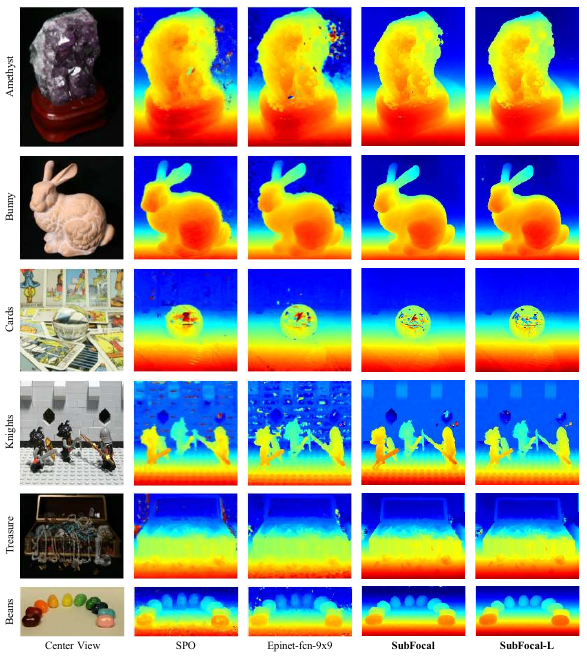}

  \caption{Visual results achieved by SPO\cite{zhang2016robust}, EPINET\cite{shin2018epinet}, and our method on the Stanford Gantry LF dataset \cite{rerabek2016new}. ground-truth disparity maps of these real-world LFs are unavailable.
  }

  \label{fig: real_1}
\end{figure*}

\begin{figure*}[htb]
  \centering
  \includegraphics[width=\linewidth]{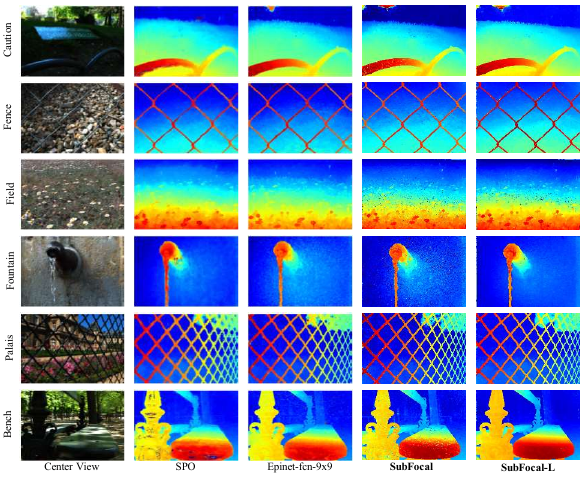}

  \caption{Visual results achieved by SPO \cite{zhang2016robust}, EPINET \cite{shin2018epinet}, and our method on LFs captured by Lytro cameras \cite{le2018light, rerabek2016new}.
  }

  \label{fig: real_2}
\end{figure*}


\subsection{Results on different LF datasets}
\label{sub:visual}

Fig. \ref{fig: real_1}, Fig. \ref{fig: real_2} compare the visual results obtained by SPO \cite{zhang2016robust}, and EPINET \cite{shin2018epinet} and our method on various types of LF datasets \cite{le2018light,rerabek2016new,vaish2008new,wanner2013datasets}.

\subsection{Future Directions}
\label{sub:future}

Although our model has shown promising performance on various datasets, there is still room for further improvement in terms of robustness and speed. In future work, we plan to enhance the robustness of our method to challenging situations, such as non-Lambertian surfaces, weak textures, and highlight regions. We will also explore the use of viewpoint selection and cost metrics to reduce the model's inference time. We believe that our method will contribute to the advancement of depth estimation based on light field imaging and promote further research in this field.

 

\bibliographystyle{IEEEtran}
\bibliography{reference}


 



